\documentclass[final,3p,times]{elsarticle}

\usepackage{amssymb}
\usepackage{amsmath}
\usepackage{xurl}
\usepackage{hyperref}
\hypersetup{hidelinks}
\usepackage{subcaption}
\usepackage{graphicx}
\usepackage{booktabs}
\usepackage{array}
\usepackage{multirow}

\newcommand{\tableformat}{\footnotesize\setlength{\tabcolsep}{2.5pt}\renewcommand{\arraystretch}{1.12}}
\newcommand{\tablehead}[1]{\shortstack{#1}}
\newcommand{\samplestat}[2]{#1 $\pm$ #2}
\newcommand{\stackstat}[3][c]{\shortstack[#1]{#2\\$\pm$ #3}}
\newcommand{\meanlabel}[1]{\raisebox{9.4pt}[0pt][0pt]{#1}}
\newcommand{\Dsim}{\mathbb{D}_{\mathrm{sim}}}
\newcommand{\Dipsc}{\mathbb{D}_{\mathrm{iPSC}}}
\newcommand{\Dmouse}{\mathbb{D}_{\mathrm{mouse}}}
\newcounter{algorithm}
\newenvironment{algorithmblock}[1]{%
\refstepcounter{algorithm}%
\par\medskip\noindent
\begin{minipage}{\linewidth}
\hrule\vspace{0.45em}
\textbf{Algorithm \thealgorithm: #1}\par
\vspace{0.35em}\hrule\vspace{0.4em}
\footnotesize
}{%
\vspace{0.35em}\hrule
\end{minipage}
\par\medskip
}
\newcommand{\algline}[2]{%
\par\noindent\hangindent=2.8em\hangafter=1%
\makebox[2.2em][r]{#1:}\hspace{0.6em}#2%
}
\ifdefined\XeTeXversion
\newcommand{\tablebest}[1]{%
  #1%
  }
\else
\newcommand{\tablebest}[1]{\textbf{#1}}
\fi

\journal{Computer Methods in Applied Mechanics and Engineering}

\begin{document}

\begin{frontmatter}

\title{MGRD: Compact morphology-gated residual diffusion for variance-aware cross-domain neurite forecasting}

\author[a]{Tsung Yeh Hsieh}
\author[a]{Cosmin Anitescu}
\author[a]{Chunghwan Kim}
\author[a]{Victoria A. Webster-Wood}
\author[a,b,c]{Yongjie Jessica Zhang}

\address[a]{Department of Mechanical Engineering, Carnegie Mellon University, 5000 Forbes Ave, Pittsburgh, PA 15213, USA}
\address[b]{Department of Biomedical Engineering, Carnegie Mellon University, 5000 Forbes Ave, Pittsburgh, PA 15213, USA}
\address[c]{Department of Civil and Environmental Engineering, Carnegie Mellon University, 5000 Forbes Ave, Pittsburgh, PA 15213, USA}

\begin{abstract}
Tracking neurite morphology over time helps characterize structural changes during neuronal development and deterioration, but long-term time-lapse imaging is resource-intensive and difficult to scale.
Forecasting future morphology could reduce this acquisition burden. The previously reported neurite digital-twin framework uses a deterministic MetaFormer-based SimVP model equipped with gated spatiotemporal attention (gSTA), yielding a single future sequence without a measure of disagreement among plausible forecasts.
This work introduces Morphology-Gated Residual Diffusion (MGRD), a compact stochastic surrogate that jointly forecasts twenty future neurite-morphology frames from ten observed frames while conditioning on a skeleton and distance field derived from the latest observation.
A two-scale morphology pathway and four spatial gates incorporate structural information into denoising, while zero-initialized projections preserve the pretrained backbone's function at the start of adaptation.
On controlled phase-field trajectories, MGRD reduces trajectory-wise mean absolute error (MAE) across ten independently sampled forecasts by 9.7\% compared with a control trained for the same adaptation period without morphology gating, while updating 4.46 times fewer parameters.
On human induced pluripotent stem cell (iPSC)-derived neuron microscopy, MGRD improves all four reported metrics relative to gSTA, including a 39.6\% reduction in trajectory-wise mean MAE and a 45.3\% increase in a skeleton-based F1 score that measures neurite centerline agreement.
Without mouse-domain retraining or fine-tuning, MGRD also improves trajectory-wise mean MAE and skeleton F1 on mouse cortical-neurosphere microscopy across 10--40-min sampling intervals, at forecast horizons extending beyond 13 hours after the last observation.
Repeated sampling yields a scalar case-level variance score that ranks cases by forecast difficulty. Retaining approximately 60\% of the lowest-variance cases reduces mean MAE relative to full coverage by 17.6\% on iPSC microscopy data and 16.8\% on simulation data.
MGRD uses 1.01\% of the parameters in gSTA and requires less than one tenth of the gSTA training-update time in either training stage. With morphology features cached, one trajectory generated with 50 DDIM steps is also 7.9\% faster than a gSTA forecast.
These results support MGRD as a compact stochastic surrogate for forecasting neurite morphology across simulation and microscopy datasets and prioritizing cases for review.
\end{abstract}


\end{frontmatter}

\section{Introduction}

Neurodevelopmental disorders encompass heterogeneous conditions in which altered neuronal development can disrupt circuit formation and function~\cite{thapar2017neurodevelopmental}. Reflecting this heterogeneity, the diagnostic concept has evolved toward a dimensional view that recognizes substantial overlap among developmental phenotypes~\cite{morris2020neurodevelopmental}. At the cellular scale, impaired dendrite morphogenesis has emerged as a recurring mechanism across several of these disorders~\cite{copf2016impairments}, while reported morphological variants involving both dendrites and axons further connect cellular structure to disorder-associated changes in neuronal connectivity~\cite{myers2019morphological}.

These dynamic structural changes can be observed directly in time-lapse neuron cultures, which capture events such as branching, extension, retraction, and fragmentation and support longitudinal morphological analysis~\cite{liao2023semi}. Quantitative analyses of these sequences span multiple temporal and spatial scales: change-point methods can identify transitions between developmental regimes~\cite{liao2021quantification}, while network-level studies reveal emergent dynamics~\cite{isomura2023experimental}. Recent cortical-neurosphere imaging further links microscopic neurite kinematics to emergent macroscopic axon topology~\cite{kim2026cortical}. However, the long, densely sampled acquisitions required for such analyses remain costly, time-intensive, and difficult to scale across the broad biological and environmental conditions considered in neuronal-culture research~\cite{soriano2023neuronal}.
A surrogate that forecasts later morphology from an early image history can reduce repeated simulation cost and help prioritize cultures for continued observation or manual review.
For this capability to be useful beyond a fixed benchmark, it must preserve sparse neurite structure, operate across microscopy settings, and indicate which forecasts warrant greater scrutiny.

Mechanistic descriptions of neuronal morphogenesis span several levels of abstraction. Early formulations linked dendritic patterns to morphological instabilities~\cite{hentschel_instabilities_1994}, whereas subsequent mathematical and biophysical models formalized axon guidance, formation, and pathfinding~\cite{krottje_mathematical_2007,pearson_mathematical_2011,aeschlimann_biophysical_2001}. At a more generative level, rule-based and context-aware approaches reproduce branching while accounting for environmental constraints~\cite{cuntz_one_2010,torben-nielsen_context-aware_2014}. These principles have also been used to generate realistic three-dimensional cortical neurons and networks in which connectivity emerges from independently growing neurites and geometry-based synapse formation~\cite{eberhard_neugen_2006,van_ooyen_independently_2014}.

For continuous representations of changing morphology, phase-field models describe axonal extension through an evolving diffuse interface~\cite{takaki_phase_field_2015}. Integrating this formulation with isogeometric analysis (IGA) supports smooth numerical representations of growing neurite geometries~\cite{qian_modeling_2022}, and later work connected biomimetic IGA simulations to morphometric features and convolutional prediction~\cite{qian2023biomimetic}.
Although these formulations provide mechanistic detail, repeatedly solving their coupled nonlinear equations over expanding, locally refined domains remain expensive when many trajectories or timely experimental decisions are required~\cite{qian2025neurodevelopmental}.
Digital twins provide a general framework for connecting a physical process to a dynamically updated computational representation~\cite{singh2021digital}. Their defining feature is the nature and direction of information exchanged with the physical counterpart~\cite{jones2020characterising}. Although digital-twin workflows first gained prominence in industrial monitoring and prediction~\cite{tao2018digital,liu2021review}, their combination with artificial intelligence increasingly supports individualized prediction and decision support in biomedical settings~\cite{rathore2021role}.

Within such a workflow, forecasting later neurite morphology from observed images becomes a structured video-prediction problem, for which recurrent rollout can accumulate error over time~\cite{oprea2020review}. Convolutional LSTMs introduced spatial recurrence for frame-sequence prediction~\cite{shi2015convolutional}, while self-attention provided an alternative for representing long-range dependencies~\cite{vaswani2017attention}. Transformer-based representations were subsequently adapted to video generation and prediction~\cite{yan2021videogpt,neimark2021video}, followed by more efficient architectures for spatiotemporal attention~\cite{ye2023video}. PredFormer further demonstrated the effectiveness of Transformers for spatiotemporal predictive learning~\cite{tang2024predformer}. More generally, MetaFormer isolates the role of the token-mixing architecture~\cite{yu2022metaformer}, and OpenSTL provides a common benchmark for comparing these predictors~\cite{tan2023openstl}. Together, these developments provide increasingly capable backbones for long-horizon spatiotemporal prediction, creating an opportunity to adapt them to domain-specific problems with additional structural and uncertainty requirements.

Qian et al.~\cite{qian2025high} developed a high-throughput machine-learning framework that predicted twenty future neurite frames from ten observed frames in synthetic and experimental datasets.
Its deterministic predictor used SimVP with a MetaFormer-style gated spatiotemporal attention (gSTA) temporal module implemented in OpenSTL~\cite{gao2022simvp,tan2022simvp,yu2022metaformer,tan2023openstl}.
The resulting gSTA model established an end-to-end connection between IGA-generated phase-field trajectories, experimental microscopy sequences, and fast deterministic spatiotemporal prediction.
Its deterministic formulation, however, returns one future sequence and does not expose disagreement among plausible forecasts.
This distinction matters in two-dimensional microscopy images, where focus changes, intensity variation, unobserved out-of-plane motion, and structures outside the field of view can affect apparent future morphology.
These limitations motivate three requirements for the forecasting framework: explicit treatment of sparse centerline structure, joint long-horizon forecasting without recursive rollout, and a case-level signal for prioritizing forecasts that may require closer review.

Diffusion probabilistic modeling formulates data generation as the learned reversal of progressive data corruption~\cite{sohl2015deep}. Denoising diffusion probabilistic models established a practical, high-quality realization of this process~\cite{ho2020denoising}, while deterministic implicit sampling reduced the number of reverse steps without requiring a different training objective~\cite{song2020denoising}. Together, these developments provide a natural basis for formulating future neurite prediction as conditional generation.
Instead of learning only a point estimate, a conditional diffusion model learns to reverse a noising process and can generate multiple future sequences for the same input history.
These samples provide both an ensemble forecast and empirical variability that can be evaluated as a case-ranking signal.
Spatial conditions can be introduced into a fitted diffusion backbone through either a trainable control branch or a lightweight structural adapter~\cite{zhang2023adding,mou2023t2i}.
Conditional video diffusion further demonstrates that future sequences can be generated jointly rather than through recursive frame-wise rollout~\cite{voleti2022mcvd}.
Morphology- and kinematics-aware conditional diffusion has also been explored for biomedical image synthesis~\cite{qi2025knowledge}.
These ideas motivate a neurite-specific design in which centerline morphology is encoded independently and injected at multiple denoiser resolutions.

Three data sources were selected to play complementary roles in the work presented here. Phase-field trajectories provide a controlled synthetic setting in which the effect of morphology-gated adaptation can be isolated; the iPSC-derived neuron dataset provides the matched experimental benchmark used in the prior gSTA study; and the separately acquired mouse cortical-neurosphere dataset provides an external domain with different cell origin, culture organization, image appearance, and temporal cadence for evaluating transfer and long-horizon robustness. This paper introduces Morphology-Gated Residual Diffusion (MGRD), a compact stochastic surrogate for morphology-aware neurite forecasting.
The new contributions of this paper are:
\begin{itemize}
    \item MGRD uses a two-scale morphology pathway and four spatial gates to preserve sparse neurite structure, while jointly denoising the complete 20-frame residual sequence without relying on recursive future-frame rollout.
    \item A two-stage adaptation procedure first fits the residual-diffusion backbone and then learns zero-initialized morphology corrections while freezing the backbone; in controlled simulation experiments, it lowers MAE relative to ungated continuation while updating 4.46 times fewer parameters.
    \item Repeated sequence sampling produces an ensemble-mean forecast, spatial variance maps, and a scalar case-level variance score that supports selective review of difficult forecasts.
    \item Experiments establish improved prediction on the matched iPSC-derived neuron microscopy benchmark, directional transfer across synthetic and experimental domains, and robustness on the external mouse cortical-neurosphere domain over 10--40-min sampling intervals without target-domain fine-tuning, while using only 1.01\% as many total parameters as gSTA (a 98.9-fold reduction, or approximately two orders of magnitude) and substantially reducing training-step latency.
\end{itemize}

The remainder of the paper is organized as follows.
Section~\ref{sec:overview} defines the long-horizon forecasting task, neurite-specific requirements, and framework outputs.
Section~\ref{sec:method} presents the morphology descriptors, residual-diffusion formulation, gated denoiser, and two-stage adaptation procedure.
Section~\ref{sec:data_evaluation} describes the simulation and microscopy datasets together with the evaluation protocol, and Section~\ref{sec:results} reports the controlled, in-domain, cross-domain, uncertainty, and efficiency results.
Section~\ref{sec:conclusion} summarizes the findings, limitations, and directions for future work.

\section{Forecasting problem and framework overview}
\label{sec:overview}

\subsection{Forecasting task and neurite-specific requirements}

The proposed framework extends the existing neurite digital-twin workflow from deterministic prediction to probabilistic forecasting.
For every sequence, the observed history and future target are defined as
\begin{equation}
    X=(X_0,\ldots,X_9), \qquad
    Y=(Y_{10},\ldots,Y_{29}),
    \label{eq:forecasting_task}
\end{equation}
where $X_j$ and $Y_j$ are grayscale frames ordered by acquisition time, and $X\in[0,1]^{T_{\mathrm{in}}\times H\times W}$ and $Y\in[0,1]^{T_{\mathrm{out}}\times H\times W}$ after preprocessing, with history length $T_{\mathrm{in}}=10$, forecast length $T_{\mathrm{out}}=20$, and image height and width $H=W=256$.
The forecasting horizon therefore spans twice as many frames as the observed history, and all twenty targets are predicted jointly rather than by recursively feeding predicted frames back into the model.
The deterministic gSTA reference produces a single estimate $\widehat{Y}_{\mathrm{gSTA}}=f_{\varphi}(X)$, where $f_\varphi$ is the predictor with learned parameters $\varphi$.
MGRD, with learned parameters $\theta$, produces multiple future trajectories from the conditional predictive distribution $p_\theta$,
\begin{equation}
    \widehat{Y}_{\mathrm{MGRD}}^{(m)} \sim
    p_{\theta}\!\left(\,\cdot\mid X,S(X_9),D(S(X_9))\right),
    \qquad m=1,\ldots,M,
    \label{eq:conditional_forecasting}
\end{equation}
where $S(X_9)$ and $D(S(X_9))$ denote the skeleton and normalized distance-transform descriptors derived from the latest observation, respectively; their construction is detailed later in Section~\ref{subsec:condition}.
Here, $M$ denotes the total number of independently generated forecasts for a given input history, and $m$ indexes one trajectory in this ensemble.
Each $\widehat{Y}_{\mathrm{MGRD}}^{(m)}$ has the same twenty-frame dimensions as $Y$; the collection $\{\widehat{Y}_{\mathrm{MGRD}}^{(m)}\}_{m=1}^{M}$ forms the trajectory ensemble.
The primary evaluations use $M=10$, as specified later in Section~\ref{subsec:evaluation}.

Long-horizon neurite forecasting presents three connected requirements.
First, the twenty-frame horizon requires temporal coherence across the full forecast, because stepwise reuse of predictions can compound errors and progressively distort fine branches.
Second, neurites occupy a sparse fraction of each image, so global image losses can be dominated by persistent background even when small centerline structures are biologically important.
The forecast must therefore retain both the centerline topology and the surrounding intensity field that describes branch width and local cellular context.
Third, microscopy histories can admit multiple plausible futures because apparent morphology is affected by intensity variation, focus changes, unobserved out-of-plane motion, and structures outside the field of view.
A useful stochastic surrogate must consequently provide not only predicted trajectories but also spatial disagreement and a scalar case-level score that can prioritize forecasts for review.

\subsection{Overview of the MGRD workflow}

\begin{figure}[ht]
\centering
\includegraphics[width=0.98\textwidth]{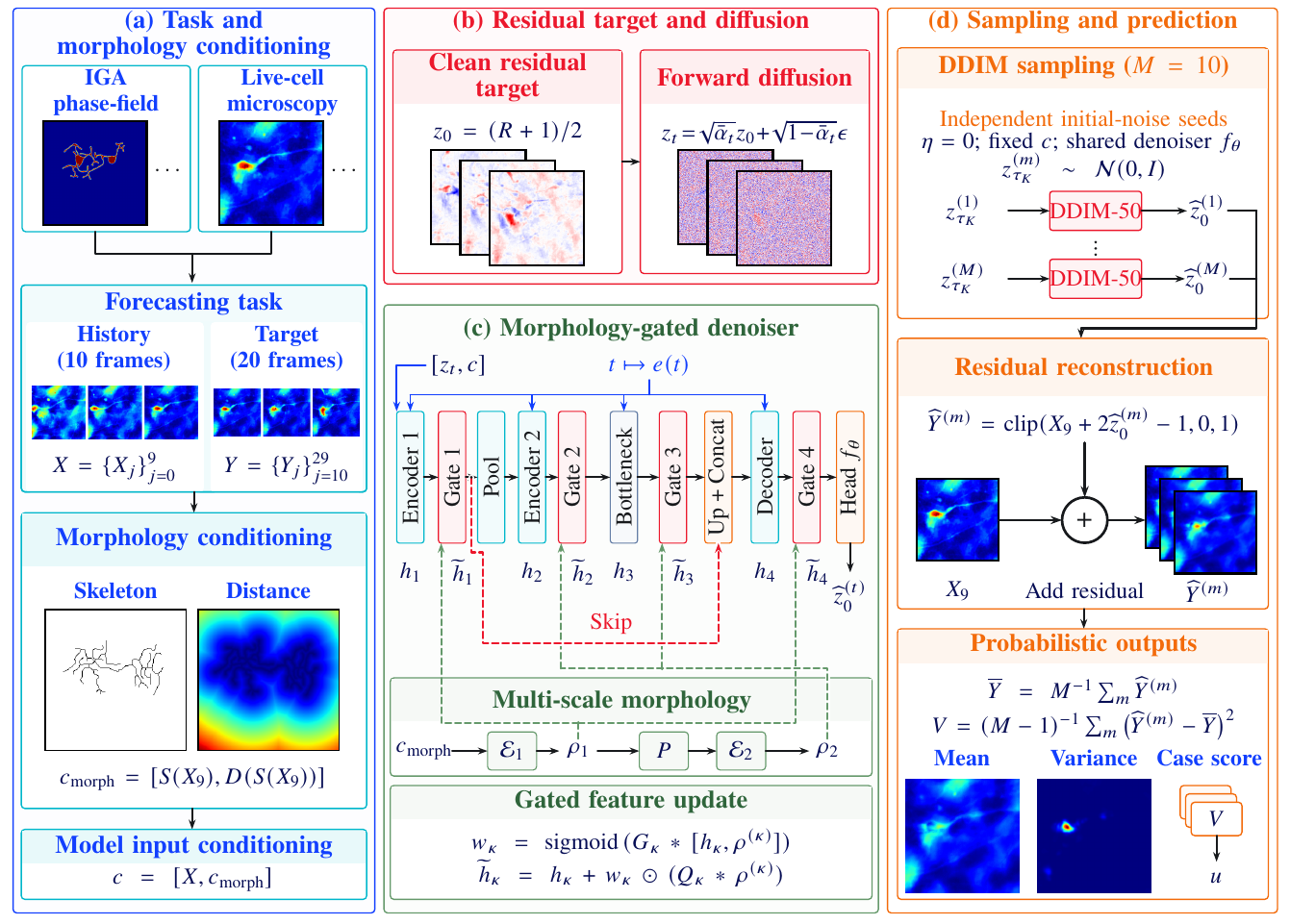}
\input{figure1_caption.tex}
\caption{Overview of MGRD for probabilistic neurite forecasting.
(a) The forecasting task and morphology conditioning derived from the observed history.
(b) Residual target encoding and forward diffusion for training.
(c) The morphology-gated denoiser, which integrates temporal context and multiscale structural features.
(d) Repeated sampling and residual reconstruction produce an ensemble of future trajectories, summarized by the ensemble mean, spatial variance, and a case-level variance score for selective prediction.}
\label{fig:framework_overview}
\end{figure}

MGRD formulates neurite forecasting as the conditional generation of future changes relative to the latest observation. The workflow illustrated in Figure~\ref{fig:framework_overview} combines temporal context with explicit morphology descriptors to generate an ensemble of possible future trajectories and summarize their variability.

The observed history provides the temporal context for forecasting, while morphology descriptors make the current neurite structure explicitly available to the model. As illustrated in Figure~\ref{fig:framework_overview}(a), the condition combines ten observed frames with a skeleton and a normalized distance transform derived from the latest observation. These descriptors represent the neurite centerlines and the surrounding spatial field, complementing the intensity information in the history, the descriptor construction is detailed in Section~\ref{subsec:condition}. Both IGA phase-field simulations and live-cell microscopy use this input representation.

To learn future changes, MGRD expresses the twenty-frame target sequence as residuals relative to the latest observation frame and rescales them into the range used by the diffusion process. The training process illustrated in Figure~\ref{fig:framework_overview}(b) shows the forward process of adding Gaussian noise to this joint residual representation. Recovering the clean target requires the model to combine information about the noisy residual, the observed history, and the diffusion step. The denoiser shown in Figure~\ref{fig:framework_overview}(c) incorporates a separate morphology pathway whose multiscale features gate corrections within the network. This connects residual recovery to the observed neurite structure while predicting the full forecast horizon jointly. Sections~\ref{subsec:residual_diffusion} and~\ref{subsec:morphology_denoiser} specify the diffusion formulation and gated architecture, respectively.

At inference, the learned denoiser generates future residual sequences from independent Gaussian noise realizations, with the observed condition and fitted model held fixed. Figure~\ref{fig:framework_overview}(d) illustrates how iterative sampling and residual reconstruction convert these realizations into an ensemble of future trajectories. Each generated residual sequence is mapped back to its original intensity range and added to the latest observation. The ensemble mean summarizes the forecast, while the sample variance measures disagreement among the generated trajectories. Averaging this variance across future frames and pixels yields a case-level score for selective prediction. The ensemble statistics and their evaluation are defined in Section~\ref{subsec:evaluation}.

\section{MGRD model formulation and training}
\label{sec:method}
This section details the construction of morphology descriptors, the residual diffusion formulation, the morphology-gated denoiser, and the two-stage adaptation procedure.

\subsection{Morphology descriptor construction}
\label{subsec:condition}

\begin{figure}[!htbp]
\centering
\includegraphics[width=\textwidth]{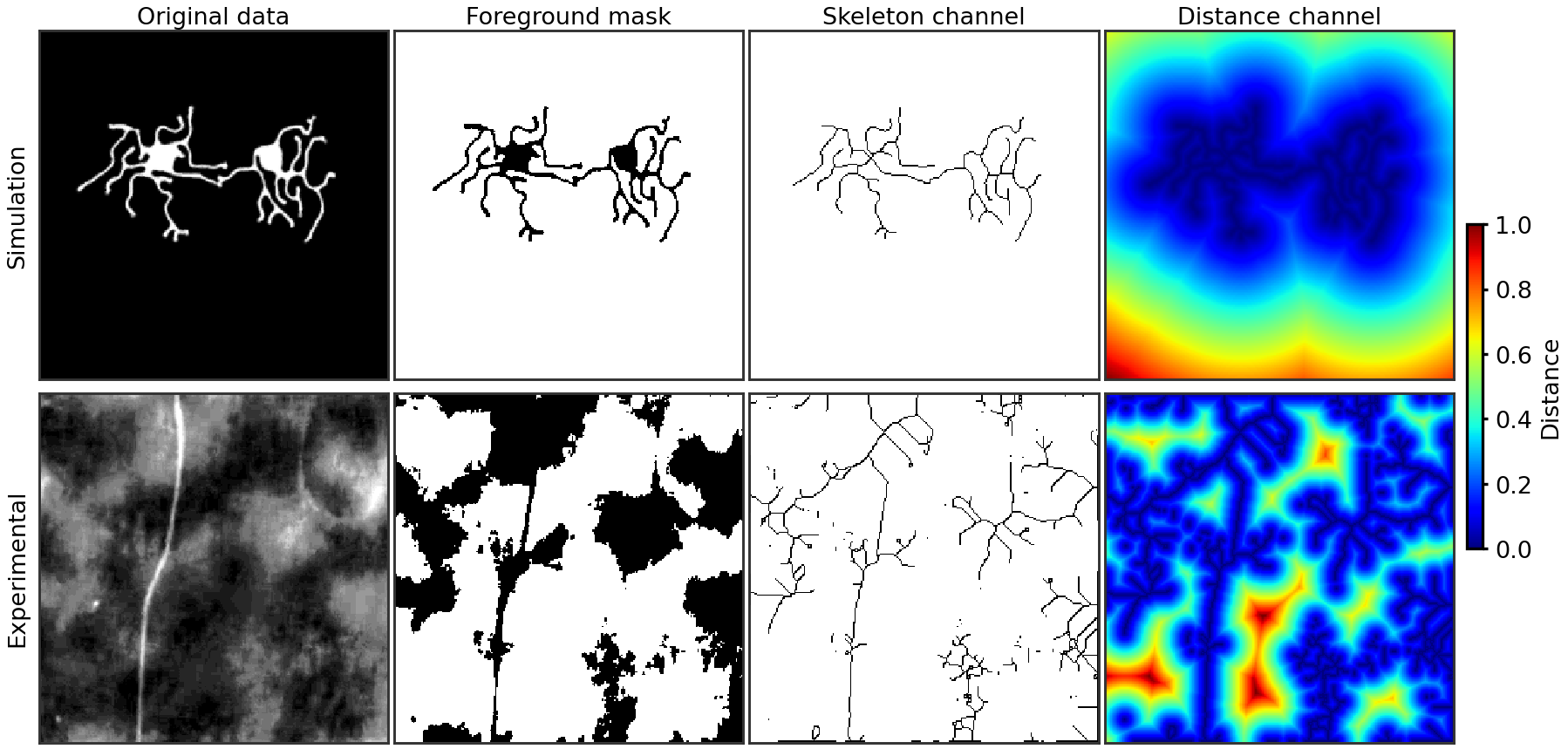}
\caption{Construction of the morphology-derived condition tensor.
The rows show the process of construction of the skeleton and distance channels from one of the representative simulation and experimental images.
Each displayed frame is thresholded to obtain a foreground mask, thinned into a neurite skeleton, and converted into a normalized distance-transform map.
The skeleton and distance channel are then concatenated with the ten observed frames to form the 12-channel condition for the denoiser.}
\label{fig:morphology_condition}
\end{figure}

MGRD uses the ten observed frames as temporal context and augments them with morphology descriptors extracted from the latest observed frame $X_9$.
The first descriptor is a skeleton image $S(X_9)$ that approximates the neurite centerline.
The second descriptor is a distance-transform image $D(S(X_9))$ that encodes spatial distance to the skeleton.
The conditional input $c$ is formed by channel
 concatenation, denoted by square brackets:
\begin{equation}
    c = \left[X_0,\ldots,X_9,S(X_9),D(S(X_9))\right].
\end{equation}
As illustrated in Figure~\ref{fig:morphology_condition}, the latest observed frame is first converted into a foreground mask and then thinned into a one-pixel-wide skeleton.
Let $I_9=\max(0,\min(X_9,1))$ denote the normalized latest frame, where the minimum and maximum operations are applied elementwise and restrict every value of $X_9$ to the interval $[0,1]$.
Let $\lambda^\star$ be the Otsu threshold that maximizes between-class variance~\cite{otsu1979threshold},
\begin{equation}
    \lambda^\star
    =
    \operatorname*{arg\,max}_{\substack{\lambda \in [0,1]\\\omega_0(\lambda)\omega_1(\lambda)>0}}
    \left\{
    \omega_0(\lambda)\omega_1(\lambda)
    \left[\mu_0(\lambda)-\mu_1(\lambda)\right]^2
    \right\} ,
\end{equation}
where $\omega_0(\lambda),\omega_1(\lambda)$ and $\mu_0(\lambda),\mu_1(\lambda)$ are the class probabilities and mean intensities for the intensity classes at or below and above a candidate threshold $\lambda$, respectively.
The optimal threshold $\lambda^\star$ is computed automatically over partitions of the 256-bin intensity histogram of $I_9$ that yield two nonempty classes.
The binary foreground mask is therefore defined explicitly as
\begin{equation}
    M_{\mathrm{fg}}(p)=
    \begin{cases}
        1, & I_9(p)>\lambda^\star,\\
        0, & I_9(p)\leq\lambda^\star,
    \end{cases}
\end{equation}
for pixel location $p$.
Let $\operatorname{ZS}$ denote the Zhang--Suen two-subiteration thinning operator, which iteratively removes boundary pixels while preserving connected centerlines~\cite{zhang1984fast}.
Applied to $M_{\mathrm{fg}}$ until convergence, it gives
\begin{equation}
    S(X_9)=\operatorname{ZS}(M_{\mathrm{fg}}).
\end{equation}
The distance-transform channel is then computed from the complement of this skeleton.
Writing $S=S(X_9)$, let $\Omega$ denote the image pixel domain and $\Omega_S=\{q\in\Omega:S(q)=1\}$ its skeleton-pixel subset.
For a nonempty skeleton and $p\in\Omega$, the Euclidean distance to the skeleton and its normalized map are
\begin{equation}
\begin{aligned}
    d(p)&=\min_{q\in\Omega_S}\|p-q\|_2,\\
    D(S)(p)&=
    \frac{d(p)-d_{\min}}
    {\max(d_{\max}-d_{\min},\varepsilon_{\mathrm{num}})},
\end{aligned}
\end{equation}
where $d_{\min}=\min_{r\in\Omega}d(r)$, $d_{\max}=\max_{r\in\Omega}d(r)$, and $\varepsilon_{\mathrm{num}}=10^{-8}$ is a numerical stabilization constant.
The skeleton and distance-transform maps each have a spatial resolution of $256\times256$ pixels.
They are concatenated with the ten observed frames along the channel axis to form a 12-channel condition for the denoiser, as shown in Figure~\ref{fig:framework_overview}(a).

\subsection{Residual diffusion formulation}
\label{subsec:residual_diffusion}

Following the conceptual formulation in Section~\ref{sec:overview}, MGRD applies diffusion in an encoded residual space defined relative to the latest observation and maps each generated residual trajectory back to the image sequence. Specifically, let $X_{\mathrm{last}}=X_9$ denote the latest observed frame, broadcast across the twenty-frame target sequence $Y$, and define the bounded residual target

\begin{equation}
    R=\max\!\left(-1,\min\!\left(Y-X_{\mathrm{last}},1\right)\right), \qquad z_0=\frac{R+1}{2}.
    \label{eq:residual_target}
\end{equation}
The diffusion process operates on the encoded residual $z_0\in[0,1]^{T_{\mathrm{out}}\times H\times W}$, with $T_{\mathrm{out}}=20$.
Following the denoising diffusion probabilistic formulation~\cite{ho2020denoising}, the forward process progressively perturbs $z_0$ with Gaussian noise over $T_d$ diffusion steps.
At diffusion step $t\in\{1,\ldots,T_d\}$, $\beta_t\in(0,1)$ specifies the variance of the added noise, while the preceding state $z_{t-1}$ is scaled by $\sqrt{1-\beta_t}$ to form the noisy state $z_t$.
The prescribed sequence $\{\beta_t\}_{t=1}^{T_d}$ forms the noise variance schedule and defines the forward Markov chain $q$, with $z_{1:T_d}=(z_1,\ldots,z_{T_d})$:
\begin{align}
    q(z_{1:T_d}\mid z_0)
    &=\prod_{t=1}^{T_d}q(z_t\mid z_{t-1}),
    \label{eq:forward_chain}\\
    q(z_t\mid z_{t-1})
    &=\mathcal{N}\!\left(
        z_t;\sqrt{\alpha_t}z_{t-1},\beta_t I
    \right),
    \qquad \alpha_t=1-\beta_t.
    \label{eq:forward_transition}
\end{align}
Here, $\mathcal{N}$ denotes a Gaussian distribution with mean $\sqrt{\alpha_t}z_{t-1}$ and covariance $\beta_t I$, where $I$ is the identity matrix.
Writing $\bar{\alpha}_t=\prod_{s=1}^{t}\alpha_s$, with $\bar\alpha_0=1$, permits direct sampling at any diffusion index:
\begin{equation}
    q(z_t\mid z_0)
    =\mathcal{N}\!\left(
        z_t;\sqrt{\bar{\alpha}_t}z_0,
        (1-\bar{\alpha}_t)I
    \right).
    \label{eq:forward_marginal}
\end{equation}
The implemented schedule uses $T_d=1,000$ steps and linearly increases $\beta_t$ from $10^{-4}$ to $2\times10^{-2}$, making the terminal state approximately standard Gaussian.
The DDPM reverse process $p_\theta(z_{0:T_d}\mid c)$ uses a conditional mean $\mu_\theta(z_t,t,c)$ and covariance $\Sigma_t$ at each step:
\begin{align}
    p_{\theta}(z_{0:T_d}\mid c)
    &=p(z_{T_d})\prod_{t=1}^{T_d}
      p_{\theta}(z_{t-1}\mid z_t,c),
    \qquad p(z_{T_d})=\mathcal{N}(z_{T_d};0,I),
    \label{eq:reverse_chain}\\
    p_{\theta}(z_{t-1}\mid z_t,c)
    &=\mathcal{N}\!\left(
        z_{t-1};\mu_{\theta}(z_t,t,c),\Sigma_t
    \right).
    \label{eq:reverse_transition}
\end{align}
The reverse-transition covariance is fixed as $\Sigma_t=\widetilde\beta_t I$, where $\widetilde\beta_t=\frac{1-\bar\alpha_{t-1}}{1-\bar\alpha_t}\beta_t$ is the forward-posterior variance; at $t=1$, this variance is zero and the final transition is deterministic.
Let $f_\theta(z_t,t,c)$ denote the conditional denoising network with trainable parameters $\theta$, which predicts the clean encoded residual $z_0$ from the noisy state $z_t$, diffusion step $t$, and condition $c$.
Let $\widehat z_0^{(t)}$ denote the clean encoded-residual estimate at denoising step $t$, obtained by clipping the network prediction to $[0,1]$.
The reverse mean is computed from this estimate as
\begin{align}
    \mu_\theta(z_t,t,c)
    &=
    \frac{\sqrt{\bar\alpha_{t-1}}\,\beta_t}{1-\bar\alpha_t}
    \widehat z_0^{(t)}
    +
    \frac{\sqrt{\alpha_t}\,(1-\bar\alpha_{t-1})}{1-\bar\alpha_t}
    z_t,
    \label{eq:reverse_mean}\\
    \widehat z_0^{(t)}
    &=\max\!\left(0,\min\!\left(f_\theta(z_t,t,c),1\right)\right).
    \label{eq:predicted_clean}
\end{align}
The observed history and morphology descriptors remain fixed throughout denoising.

The denoiser is trained to recover the clean encoded residual $z_0$ directly.
For each training pair $(z_0,c)$, the diffusion index $t$ is sampled uniformly from $\{1,\ldots,T_d\}$ and $\epsilon\sim\mathcal{N}(0,I)$ is used to construct
\begin{equation}
    z_t=\sqrt{\bar{\alpha}_t}z_0
    +\sqrt{1-\bar{\alpha}_t}\,\epsilon.
    \label{eq:noisy_residual_sample}
\end{equation}
Within each training example, all twenty frame channels share the sampled diffusion index $t$, while the entries of $\epsilon$ are independent across channels and pixels. The stacked thumbnails in Figure~\ref{fig:framework_overview}(b) illustrate selected future-frame channels at one diffusion index, not successive diffusion steps.
The clean-state prediction loss, denoted $\mathcal{L}_{x_0}(\theta)$ with $z_0$ as the clean target, is the elementwise mean squared error
\begin{equation}
    \mathcal{L}_{x_0}(\theta)
    = \mathbb{E}_{(z_0,c),t,\epsilon}
    \left[
    \frac{1}{T_{\mathrm{out}}HW}
    \left\|f_{\theta}(z_t,t,c)-z_0\right\|_F^2
    \right],
    \label{eq:x0_loss}
\end{equation}
where $\mathbb{E}$ averages over training pairs $(z_0,c)$, diffusion indices $t$, and Gaussian noise $\epsilon$, and $\|\cdot\|_F$ denotes the Frobenius norm over the twenty output channels and all spatial locations.
In practice, the expectation is estimated by averaging the per-sample loss over each mini-batch.
At inference, $\widehat z_0^{(t)}$ is converted to the implied noise estimate $\widehat\epsilon_t$:
\begin{equation}
    \widehat\epsilon_t
    =\frac{z_t-\sqrt{\bar\alpha_t}\,\widehat z_0^{(t)}}
    {\sqrt{1-\bar\alpha_t}}.
    \label{eq:implied_noise}
\end{equation}
The main experiments use $K=50$ denoising diffusion implicit model (DDIM) updates at selected diffusion indices $\tau_K>\cdots>\tau_1$, where $k$ indexes the selected steps, $\tau_K=T_d$, and $\tau_0=0$ denotes the clean endpoint with $\bar\alpha_{\tau_0}=1$.
In the DDIM sampler~\cite{song2020denoising}, $\eta\in[0,1]$ controls the injected noise, $\sigma_k$ is its standard deviation at update $k$, and $\xi_k$ is an independent standard Gaussian draw.
The update from $\tau_k$ to $\tau_{k-1}$ is
\begin{align}
    \sigma_k
    &=\eta\sqrt{
        \frac{1-\bar\alpha_{\tau_{k-1}}}
             {1-\bar\alpha_{\tau_k}}
        \left(1-
        \frac{\bar\alpha_{\tau_k}}
             {\bar\alpha_{\tau_{k-1}}}
        \right)},
    \label{eq:ddim_sigma}\\
    z_{\tau_{k-1}}
    &=\sqrt{\bar\alpha_{\tau_{k-1}}}\,\widehat z_0^{(\tau_k)}
      +\sqrt{1-\bar\alpha_{\tau_{k-1}}-\sigma_k^2}\,
       \widehat\epsilon_{\tau_k}
      +\sigma_k\xi_k,
    \qquad \xi_k\sim\mathcal{N}(0,I).
    \label{eq:ddim_update}
\end{align}
In this work, we use $\eta=0$, which gives a simplified form:
\begin{align}
    z_{\tau_{k-1}}
    &=\sqrt{\bar\alpha_{\tau_{k-1}}}\,\widehat z_0^{(\tau_k)}
      +\sqrt{1-\bar\alpha_{\tau_{k-1}}}\,
       \widehat\epsilon_{\tau_k}.
    \label{eq:ddim_update_eta0}
\end{align}

Therefore, each trajectory is deterministic conditional on its initial draw $z_{\tau_K}\sim\mathcal{N}(0,I)$; independent initial-noise seeds produce the repeated forecasts used for the ensemble mean and sample variance.
Let $\widehat z_0$ denote the final encoded residual at $\tau_0=0$.
The predicted image sequence $\widehat Y$ is reconstructed as
\begin{equation}
    \widehat{Y}=\max\!\left(0,\min\!\left(X_{\mathrm{last}}+2\widehat{z}_0-1,1\right)\right).
    \label{eq:residual_reconstruction}
\end{equation}

\subsection{Morphology-gated denoiser}
\label{subsec:morphology_denoiser}

\begin{figure*}[!ht]
\centering
\includegraphics[width=1.0\textwidth]{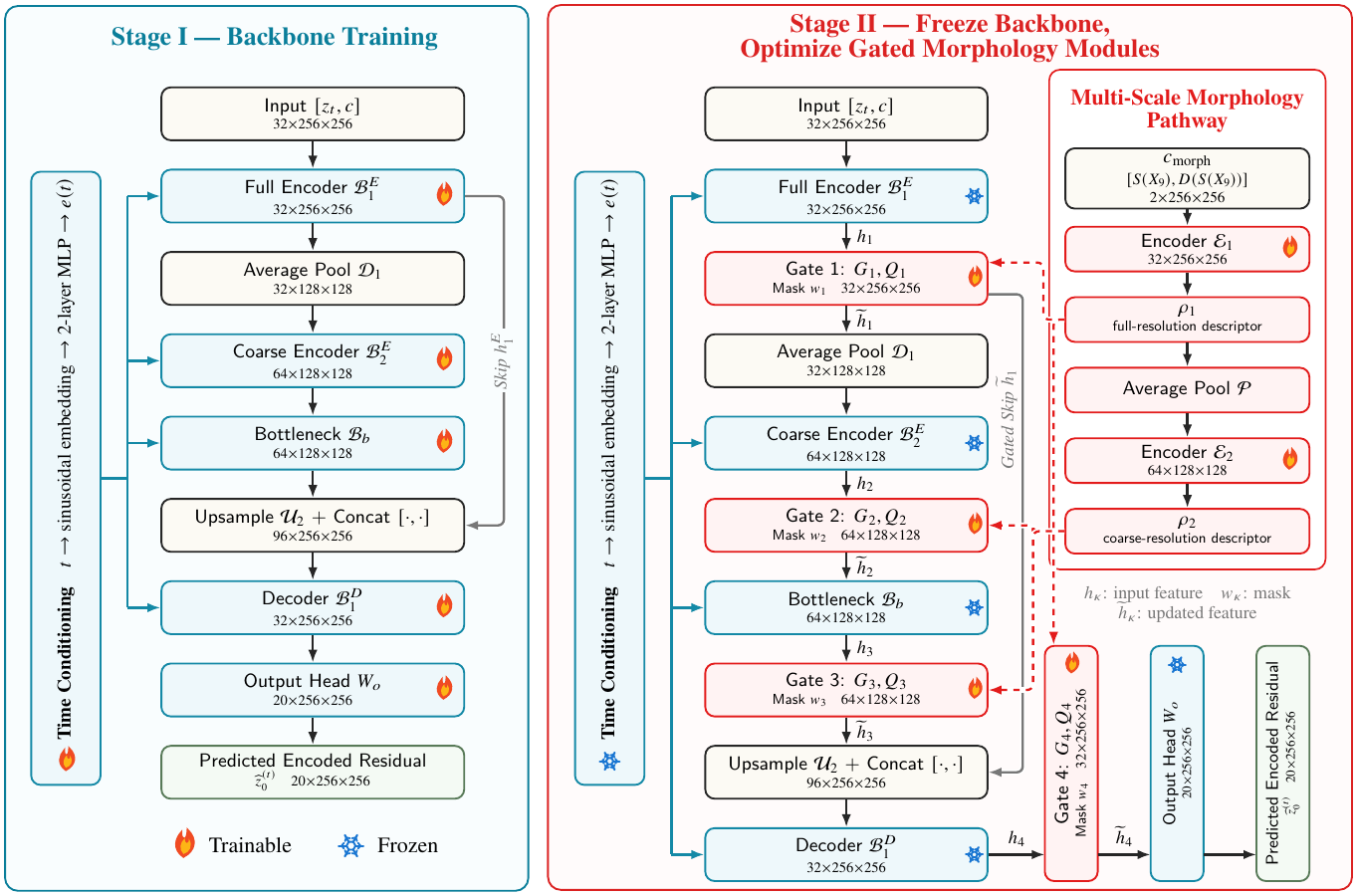}
\caption{MGRD architecture and two-stage optimization.
The teal backbone receives $[z_t,c]$ and injects the timestep embedding $e(t)$ into each convolutional block; the red morphology pathway encodes the morphology input $c_{\mathrm{morph}}$ into a full-resolution descriptor $\rho_1$ and a coarse-resolution descriptor $\rho_2$.
Gates 1--4 operate after the full-resolution encoder, coarse encoder, bottleneck, and decoder, respectively, using morphology features $\rho_1$, $\rho_2$, $\rho_2$, and $\rho_1$; the gated output of Gate~1 also forms the full-resolution skip connection to the decoder.
Each gate is labeled with its convolutions $G_\kappa,Q_\kappa$ and mask $w_\kappa$; the adjoining arrows identify the incoming feature $h_\kappa$ and updated feature $\widetilde h_\kappa$.
Zero-initialized morphology projections ($Q_\kappa^{(0)}=0$) preserve the Stage I function when the Stage II adapters, parameterized by $\theta_{\mathrm{morph}}$, are inserted.
Both pipelines end with the predicted encoded residual $\widehat z_0^{(t)}$, obtained from the output head $W_o$ followed by clipping to $[0,1]$.}
\label{fig:conditional_unet}
\end{figure*}

The skeleton and distance-transform channels expose the sparse centerline geometry and its spatial neighborhood through both the early-concatenated condition $c$ and a dedicated multi-scale morphology pathway, as outlined in Figure~\ref{fig:framework_overview}(c).
Let $\theta_{\mathrm{backbone}}$ and $\theta_{\mathrm{morph}}$ denote the parameters of the denoising backbone and morphology adapters, respectively, with $\theta=(\theta_{\mathrm{backbone}},\theta_{\mathrm{morph}})$ for the complete model.
Figure~\ref{fig:conditional_unet} summarizes this dual conditioning, timestep injection into each backbone block, the gated skip connection, and the four morphology-injection points.
The multi-resolution gates allow the centerline condition to correct denoising features where its spatial information is useful, rather than restricting morphology to a single input-level fusion.

The denoising network follows the multiresolution encoder--decoder construction introduced by U-Net~\cite{ronneberger2015unet}, a structure that has also served as the denoiser for diffusion-based field prediction in computational mechanics~\cite{jadhav2023stressd}.
Let $e(t)$ be the sinusoidal diffusion-timestep embedding after a two-layer multilayer perceptron.
For an input feature map $h$ and embedding $e=e(t)$, convolutional block $\ell$ is
\begin{equation}
    \mathcal{B}_{\ell}(h,e)
    = \operatorname{SiLU}\!\left(
    W_{\ell,2} *
    \left[\operatorname{SiLU}(W_{\ell,1}*h)+P_{\ell}e\right]
    \right),
    \label{eq:unet_block}
\end{equation}
where $W_{\ell,1}$ and $W_{\ell,2}$ are convolution kernels, $P_{\ell}$ projects the timestep embedding to the block width, and $*$ denotes convolution.
The projected embedding is broadcast over the spatial dimensions; biases in convolutions and linear projections are implicit in the notation throughout this subsection.
For a conditional U-Net backbone with $N_{\mathrm{res}}$ resolution levels, let $h_0^{E}=[z_t,c]$ denote the input and $\mathcal{D}_{\ell}$ and $\mathcal{U}_{\ell}$ denote downsampling and upsampling operators, respectively.
The encoder blocks $\mathcal{B}_{\ell}^{E}$ produce features $h_{\ell}^{E}$, and the decoder blocks $\mathcal{B}_{\ell}^{D}$ produce features $h_{\ell}^{D}$.
Between them, the bottleneck block $\mathcal{B}_b$ processes features at the lowest spatial resolution, mapping $h_{N_{\mathrm{res}}}^{E}$ to $h_{N_{\mathrm{res}}}^{D}$.
The output convolution $W_o$ maps the full-resolution decoder feature $h_1^{D}$ to the backbone prediction $f_{\theta_{\mathrm{backbone}}}(z_t,t,c)$.
These operations are written as
\begin{equation}
\begin{aligned}
    h_1^{E}
        &=\mathcal{B}_1^{E}(h_0^{E},e(t)), \\
    h_{\ell}^{E}
        &=\mathcal{B}_{\ell}^{E}
        \!\left(\mathcal{D}_{\ell-1}(h_{\ell-1}^{E}),e(t)\right),
        &&\ell=2,\ldots,N_{\mathrm{res}},\\
    h_{N_{\mathrm{res}}}^{D}
        &=\mathcal{B}_{b}(h_{N_{\mathrm{res}}}^{E},e(t)), \\
    h_{\ell}^{D}
        &=\mathcal{B}_{\ell}^{D}
        \!\left([\mathcal{U}_{\ell+1}(h_{\ell+1}^{D}),h_{\ell}^{E}],e(t)\right),
        &&\ell=N_{\mathrm{res}}-1,\ldots,1,\\
    f_{\theta_{\mathrm{backbone}}}(z_t,t,c)
        &=W_o*h_1^{D}.
\end{aligned}
\label{eq:conditional_unet}
\end{equation}
Skip connections concatenate encoder features $h_\ell^{E}$ with upsampled decoder features at the matching resolution.
In the implemented two-resolution network ($N_{\mathrm{res}}=2$), the full-resolution skip is taken after Gate~1, so the decoder concatenates the upsampled bottleneck feature with the gated encoder feature.

MGRD augments this backbone with an independent morphology encoder.
Let $c_{\mathrm{morph}}=[S(X_9),D(S(X_9))]$ and let $\mathcal{P}$ denote average pooling.
The full- and coarse-resolution morphology features are
\begin{equation}
    \rho_1=\mathcal{E}_1(c_{\mathrm{morph}}), \qquad
    \rho_2=\mathcal{E}_2\!\left(\mathcal{P}(\rho_1)\right),
    \label{eq:morph_encoder}
\end{equation}
where $\mathcal{E}_1$ contains two $3\times3$ convolutions and $\mathcal{E}_2$ contains one $3\times3$ convolution, each followed by SiLU activation.
At Gate $\kappa\in\{1,2,3,4\}$ in Figure~\ref{fig:conditional_unet}, let $h_\kappa$ denote the incoming backbone feature and $\rho^{(\kappa)}$ the morphology feature selected for that location, with $(\rho^{(1)},\rho^{(2)},\rho^{(3)},\rho^{(4)})=(\rho_1,\rho_2,\rho_2,\rho_1)$.
The gating convolutions $G_\kappa$ and $Q_\kappa$ produce the sigmoid mask $w_\kappa$ and the projected morphology correction, yielding the updated feature $\widetilde h_\kappa$:
\begin{align}
    w_\kappa &= \operatorname{sigmoid}\!\left(
        G_\kappa * [h_\kappa,\rho^{(\kappa)}]
    \right), \\
    \widetilde h_\kappa &= h_\kappa
        +w_\kappa\odot\left(Q_\kappa*\rho^{(\kappa)}\right),
    \label{eq:morph_gate}
\end{align}
where $G_\kappa$ and $Q_\kappa$ are pointwise $1\times1$ convolutions, $\operatorname{sigmoid}$ is the logistic sigmoid, and $\odot$ denotes elementwise multiplication.
The mask has the same channel and spatial dimensions as $h_\kappa$; each updated feature $\widetilde h_\kappa$ is passed to the next backbone operation, with Gate~1 also providing the skip feature.
Unlike early fusion alone, Eq.~\eqref{eq:morph_gate} lets the model select both where and at which resolution the centerline descriptors modify the denoising features.
The frozen-backbone control branch in ControlNet motivates the use of function-preserving initialization~\cite{zhang2023adding}, while lightweight structural adapters provide a related precedent for learning new conditions without refitting the full backbone~\cite{mou2023t2i}. The present gates and morphology hierarchy are specialized for sparse neurite forecasting.

\subsection{Two-stage adaptation}
\label{subsec:adaptation}

Figure~\ref{fig:conditional_unet} also summarizes the two-stage adaptation strategy.
Stage I first fits the early-fusion residual-diffusion backbone using the complete condition $c$.
Stage II then inserts the dedicated morphology encoder and four gated correction modules, freezes the fitted backbone, and optimizes only the morphology-adapter parameters $\theta_{\mathrm{morph}}$.
To ensure that inserting these modules does not initially perturb the fitted denoising function, the weights and biases of each morphology projection $Q_\kappa$ are initialized to zero.
Let $\theta_{\mathrm{morph}}^{(0)}$ denote the adapter parameters at this initialization.
The resulting function-preserving property is
\begin{equation}
    \widetilde h_\kappa^{(0)}=h_\kappa,\qquad
    f_{\theta_{\mathrm{backbone}},\theta_{\mathrm{morph}}^{(0)}}(z_t,t,c)
    =f_{\theta_{\mathrm{backbone}}}(z_t,t,c).
    \label{eq:function_preserving}
\end{equation}
Both stages minimize the loss $\mathcal{L}_{x_0}(\theta)$ defined in Eq.~\eqref{eq:x0_loss}: Stage I uses $\theta=\theta_{\mathrm{backbone}}$, while Stage II uses $\theta=(\theta_{\mathrm{backbone}},\theta_{\mathrm{morph}})$ with the fitted backbone parameters held fixed.
Superscript $*$ denotes the fitted parameters from each stage, whose optimization objectives are
\begin{align}
    \theta_{\mathrm{backbone}}^*
    &=\arg\min_{\theta_{\mathrm{backbone}}}\mathcal{L}_{x_0}(\theta_{\mathrm{backbone}}),
    \label{eq:stage_i}\\
    \theta_{\mathrm{morph}}^*
    &=\arg\min_{\theta_{\mathrm{morph}}}\mathcal{L}_{x_0}\!\left((\theta_{\mathrm{backbone}}^*,\theta_{\mathrm{morph}})\right),
    \qquad \theta_{\mathrm{backbone}}^*\ \text{fixed}.
    \label{eq:stage_ii}
\end{align}
Algorithm~\ref{alg:mgrd} summarizes residual encoding, forward corruption, two-stage optimization, and DDIM ensemble forecasting, following the workflow in Figure~\ref{fig:framework_overview}(d) and Figure~\ref{fig:conditional_unet}.

\begin{algorithmblock}{MGRD training and DDIM ensemble forecasting}
\label{alg:mgrd}
\textbf{Input:} Training pairs $\mathcal{T}=\{(X,Y)\}$; observed history $X$ for forecasting; schedule $\{\beta_t\}_{t=1}^{T_d}$; DDIM indices $\tau_K>\cdots>\tau_1$ with $\bar\alpha_{\tau_0}=1$; forecast samples $M$.\par
\textbf{Output:} Fitted parameters $(\theta_{\mathrm{backbone}}^*,\theta_{\mathrm{morph}}^*)$ and forecasts $\{\widehat Y^{(m)}\}_{m=1}^{M}$ with their pixelwise mean and sample variance.\par
\algline{1}{\textbf{Stage I: residual-diffusion backbone}}
\algline{2}{\textbf{for} each mini-batch $(X,Y)$ across the training epochs \textbf{do}}
\algline{3}{\hspace*{1em}Set $X_{\mathrm{last}}=X_9$; construct $c_{\mathrm{morph}}=[S(X_9),D(S(X_9))]$, $c=[X,c_{\mathrm{morph}}]$, and $z_0$ using Eq.~\eqref{eq:residual_target}.}
\algline{4}{\hspace*{1em}Sample $t\sim\mathcal{U}\{1,\ldots,T_d\}$ and $\epsilon\sim\mathcal{N}(0,I)$; form $z_t$ using Eq.~\eqref{eq:noisy_residual_sample}.}
\algline{5}{\hspace*{1em}Update $\theta_{\mathrm{backbone}}$ by minimizing $\mathcal{L}_{x_0}(\theta_{\mathrm{backbone}})$ as in Eq.~\eqref{eq:stage_i}.}
\algline{6}{\textbf{end for}; retain fitted parameters $\theta_{\mathrm{backbone}}^*$.}
\algline{7}{\textbf{Stage II: morphology adapters}}
\algline{8}{Insert the morphology encoder $(\mathcal{E}_1,\mathcal{E}_2)$ and four gates with $Q_\kappa^{(0)}=0$; freeze $\theta_{\mathrm{backbone}}^*$.}
\algline{9}{\textbf{for} each mini-batch $(X,Y)$ across the training epochs \textbf{do}}
\algline{10}{\hspace*{1em}Construct $(z_0,c)$, sample $(t,\epsilon)$, and compute $z_t$ as in lines 3--4.}
\algline{11}{\hspace*{1em}Update only $\theta_{\mathrm{morph}}$ by minimizing $\mathcal{L}_{x_0}\!\left((\theta_{\mathrm{backbone}}^*,\theta_{\mathrm{morph}})\right)$ as in Eq.~\eqref{eq:stage_ii}.}
\algline{12}{\textbf{end for}; retain fitted parameters $\theta_{\mathrm{morph}}^*$.}
\algline{13}{\textbf{Forecasting:} use $\theta=(\theta_{\mathrm{backbone}}^*,\theta_{\mathrm{morph}}^*)$; set $X_{\mathrm{last}}=X_9$ and construct $c=[X,S(X_9),D(S(X_9))]$.}
\algline{14}{\textbf{for} $m=1,\ldots,M$ \textbf{do}; draw $z_{\tau_K}^{(m)}\sim\mathcal{N}(0,I)$.}
\algline{15}{\hspace*{1em}\textbf{for} $k=K,\ldots,1$ \textbf{do}}
\algline{16}{\hspace*{2em}For sample $m$, predict $\widehat z_0^{(\tau_k)}$ and $\widehat\epsilon_{\tau_k}$ using Eqs.~\eqref{eq:predicted_clean} and~\eqref{eq:implied_noise}.}
\algline{17}{\hspace*{2em}Set $z_{\tau_{k-1}}^{(m)}$ using Eq.~\eqref{eq:ddim_update_eta0}.}
\algline{18}{\hspace*{1em}\textbf{end for}; reconstruct $\widehat Y^{(m)}$ using Eq.~\eqref{eq:residual_reconstruction}.}
\algline{19}{\textbf{end for}; \textbf{return} the trajectory ensemble $\{\widehat Y^{(m)}\}_{m=1}^{M}$, its pixelwise mean, and sample variance.}
\end{algorithmblock}

\subsection{Implementation and comparison setup}
\label{subsec:implementation}

The implemented two-level U-Net uses channel widths of 32 and 64.
Each backbone block comprises two $3\times3$ convolutions with SiLU activations and an additive projected timestep embedding; average pooling and nearest-neighbor interpolation provide down- and upsampling, respectively.
The morphology encoder produces 32-channel full-resolution features and 64-channel coarse-resolution features.
At Stage II initialization, the gate convolutions have zero weights and biases of $-2$, producing a uniform initial mask of $\operatorname{sigmoid}(-2)\approx0.119$, while the morphology projections have zero weights and biases.
The zero projections preserve the fitted backbone function exactly, whereas the negative gate bias favors a conservative but nonsaturated gate activation once the projections begin to learn.
The input layer receives 32 concatenated channels with 20 noisy residual channels, 10 history channels, and two morphology channels. The final pointwise $1\times1$ convolution maps the decoder features to 20 clean residual channels at $256\times256$ resolution.
The full model has 297,700 parameters, of which 54,544 morphology-pathway parameters are optimized in Stage II.

All training is performed using AdamW with learning rate $10^{-4}$, weight decay $10^{-2}$, batch size 2, and the 1,000-step linear diffusion schedule.
Stage I runs for 225 epochs.
Stage II uses five epochs unless stated otherwise.
The deterministic reference is the gSTA predictor introduced in the prior high-throughput neurite framework~\cite{qian2025high}. It follows the SimVP design, which separates spatial and temporal processing in a compact spatiotemporal predictive-learning framework~\cite{gao2022simvp,tan2022simvp}. Its temporal translator uses the MetaFormer-style gSTA block implemented in the OpenSTL benchmark~\cite{tan2023openstl}; this reference is denoted gSTA throughout.
gSTA and MGRD are evaluated on identical test splits using the forecasting task in Eq.~\eqref{eq:forecasting_task} and the framewise min--max preprocessing will be introduced in Section~\ref{subsec:data}.

\section{Data sources and evaluation protocol}
\label{sec:data_evaluation}

\subsection{Datasets and preprocessing}
\label{subsec:data}

This study uses three datasets with complementary experimental roles.
The IGA phase-field simulation and human induced pluripotent stem cell (iPSC)-derived neuron microscopy datasets were introduced in the prior gSTA study~\cite{qian2025high}.
The phase-field trajectories provide a controlled synthetic domain for isolating morphology-gated adaptation and evaluating transfer between simulated and experimental data.
The iPSC microscopy dataset provides the primary matched experimental benchmark and the source-trained models used for external long-horizon evaluation.
A separate mouse cortical-neurosphere microscopy collection, described in the companion tracking study~\cite{kim2026cortical}, serves as both a source and a target in the directional-transfer matrix and as the external target for evaluating iPSC-trained models across longer sampling intervals without mouse-domain fine-tuning.

The iPSC benchmark combines microscopy sequences from two neuron--astrocyte culture studies investigating complementary mechanisms of neuronal dysregulation.
One study examined how the APOE $\varepsilon$4/$\varepsilon$4 genotype diminishes astrocytic neurotrophic function~\cite{zhao2017apoe}, whereas the other investigated mitochondrial lipid-metabolism changes associated with ABCA7 deficiency~\cite{kawatani2023abca7}.
Following the prior preprocessing pipeline~\cite{qian2025high}, the source videos are segmented temporally and divided spatially into a $4 \times 4$ grid of $256 \times 256$-pixel crops.
Each resulting sequence is organized according to the forecasting task in Eq.~\eqref{eq:forecasting_task}.
The resulting iPSC-derived neuron microscopy dataset contains 403 training, 86 validation, and 87 test sequences.
Following the prior gSTA preprocessing pipeline, the input and target frames use framewise min--max normalization.
Each frame is independently mapped to $[0,1]$.
For a source frame $I_j$, the transform is
\begin{equation}
    \widetilde I_j =
    \frac{I_j-\min(I_j)}
    {\max(I_j)-\min(I_j)+\varepsilon_{\mathrm{num}}},
\end{equation}
where the same $\varepsilon_{\mathrm{num}}=10^{-8}$ also maps constant frames to zero.

The simulation dataset reuses 134 synthetic neurite deterioration trajectories prepared for the prior gSTA study using the IGA-based phase-field model described in Section~\ref{sec:iga_review}~\cite{qian2025high,qian2025neurodevelopmental}.
These trajectories use the optimal-neurotrophin-concentration perturbation protocol.
Each simulation first evolves under the healthy-growth condition $c_{\mathrm{opti}}=1$, allowing single or multiple neurons to establish extended neurite morphologies, where $c_{\mathrm{opti}}$ denotes the optimal neurotrophin concentration.
After this initial growth phase, $c_{\mathrm{opti}}$ is reduced to zero.
Because the phase-field driving force depends on the difference $c_{\mathrm{opti}}-c_{\mathrm{neur}}$, this reduction promotes inward motion of the phase-field interface when the local neurotrophin concentration exceeds the lowered optimum.
The resulting trajectories exhibit progressive neurite retraction, thinning, and atrophy, with some receding processes leaving small disconnected neurite remnants.
Each simulation sequence is mapped from unstructured solver output to a structured image grid by querying three neighboring solver points with a cKD-tree and is subsequently resized by bilinear interpolation~\cite{bentley1975multidimensional}.
To accommodate small deviations outside the nominal phase-field range and match the gSTA preprocessing domain, the simulation frames use the same framewise min--max transform.
The simulation split contains 93 training, 20 validation, and 21 test sequences.

The mouse cortical-neurosphere microscopy dataset contains 24 inverted bright-field video exports from 12 cortical neurospheres, with two acquisition phases per specimen~\cite{kim2026cortical}. Neurospheres were formed from primary embryonic day 18 cortical neurons ($\sim$7,300 cells per aggregate) and transferred on day \textit{in vitro} 6 to functionalized substrate (Poly-D-Lysine or Poly-D-Lysine with Laminin) in 24-well plates. Cultures were imaged at 20$\times$ magnification with a \(3\times3\) tiled field acquired per well at a spatial scale of \(0.35~\mu\mathrm{m\,pixel}^{-1}\) and a temporal resolution of one frame every 10 min. After a 12-hour attachment period, each specimen underwent two 10-hour continuous live-imaging sessions (Phase 1 and Phase 2) separated by a 12-hour rest in the standard incubator, with environmental conditions (37\,$^{\circ}$C, 5\% CO$_2$, controlled humidity) maintained throughout by an Okolab H301-K-FRAME stage-top incubator. More details are provided in \cite{kim2026cortical}.

Each video was converted to the model domain by extracting the green channel, applying a Sato dark-ridge filter with scales 1--3, and independently mapping every \(256\times256\) crop and frame to \([0,1]\).
Each crop therefore covers \(89.6\times89.6~\mu\mathrm{m}^{2}\).
For the native 10-min transfer experiment, each video was divided into nonoverlapping 30-frame sequences comprising 10 input and 20 target frames.
For each sequence, eight spatial crops were selected according to the neurite-like ridge signal in the final input frame, and the same crop locations were applied to all 30 frames.
The resulting 384 sequences are divided at the specimen level into 256 training, 64 validation, and 64 test sequences from eight, two, and two specimens, respectively.
For the directional-transfer evaluation reported in Section~\ref{subsec:cross_domain}, gSTA and MGRD are fitted separately on each source-domain training split and evaluated without target-domain adaptation on the fixed test split of each destination domain.
MGRD uses the same fixed five-epoch Stage-II adaptation budget for every source domain.
The transfer comparison reports deterministic gSTA and trajectory-wise mean performance across ten DDIM-50 MGRD forecasts.
The longer-horizon evaluation uses four long-duration videos from two specimens in the same mouse collection and eight fixed radial crops defined from the observed frames around each central neurosphere.
To evaluate longer source-frame horizons, a temporal stride \(s\) is applied before preprocessing:
\begin{equation}
    \mathcal{V}_{j,s}
    = \{I_j,I_{j+s},\ldots,I_{j+29s}\},
\end{equation}
with \(s\in\{1,2,4\}\).
The resulting sampling intervals are 10, 20, and 40 min.
At stride 4, the 30-frame sequence spans 1,160 min (19 h 20 min), and the latest observation and final target are separated by 800 min (13 h 20 min).
This temporal-stride evaluation applies the fitted iPSC gSTA and MGRD models directly to the cortical-neurosphere videos without mouse-domain fine-tuning.
\subsection{Controlled IGA phase-field trajectories}
\label{sec:iga_review}

To complement experimental microscopy with controlled deterioration trajectories, this study adopts the established IGA-based phase-field model and discretization from the prior framework~\cite{qian2025neurodevelopmental}.
This established mechanics model supplies the controlled synthetic domain used to evaluate morphology forecasting under physically defined deterioration dynamics.
IGA uses the same spline basis for geometric representation and numerical analysis, making it well suited to smooth biomedical domains~\cite{hughes_isogeometric_2005}. Image-based geometric modeling provides a route from scanned biological structures to analysis-ready computational domains~\cite{zhang_geometric_2018}, as illustrated by patient-specific NURBS representations developed for vascular-flow analysis~\cite{zhang2007patient}.

Local refinement is particularly important for evolving neurite geometries because it concentrates resolution near fine morphological features. Truncated T-splines provide this capability while retaining analysis-suitable basis properties~\cite{wei2017truncated}, and unstructured extensions accommodate configurations with multiple extraordinary points per face~\cite{wei2022analysis}. Related hierarchical Catmull--Clark constructions support localized refinement on complex surfaces~\cite{wei2015truncated,wei2016extended}, while truncated T-splines can also reparameterize existing NURBS surfaces without globally increasing resolution~\cite{liunurbs}.

Within this geometric setting, phase-field methods represent evolving boundaries implicitly and have been applied to both axonal extension and dendrite growth~\cite{takaki_phase_field_2015,takaki2014phase}. Because repeatedly solving the resulting systems remains costly, several surrogate strategies have been explored. Convolutional models have predicted reaction--diffusion fields~\cite{li_reaction_2020} and material transport through complex neurite networks~\cite{li_deep_2021}. PDE-constrained optimization has provided a complementary route for studying neuronal transport regulation and traffic jams in both two- and three-dimensional geometries~\cite{li2022modeling,li2022modeling3D}. More recent reduced-order approaches include an IGA-based physics-informed graph neural network~\cite{li2023isogeometric} and graph-autoencoder latent dynamics for neurite material transport~\cite{hsieh2025galds}.

The phase-field variable $\phi$ describes the neurite morphology, with values near one in the neuronal domain and near zero outside it, and $t_{\mathrm{phys}}$ denotes physical time in this subsection.
Its evolution is driven by anisotropy, local orientation, and a driving force term:
\begin{equation}
    \frac{\partial \phi}{\partial t_{\mathrm{phys}}}
    = M_{\phi} \left[
    \mathcal{A}_{\Psi}[\phi]
    + \phi(1-\phi) \left(\phi - \frac{1}{2} + F_{driv} + 6H_{\phi} |\nabla \vartheta| \right)
    \right],
    \label{eq:phase_field}
\end{equation}
where the anisotropic interfacial operator is
\begin{equation*}
    \mathcal{A}_{\Psi}[\phi]
    =-\frac{\delta}{\delta\phi}
    \int_{\Omega}\frac{1}{2}a(\Psi)^2|\nabla\phi|^2\,d\Omega.
\end{equation*}
This variational notation compactly retains the orientation-dependent correction terms generated by the anisotropic gradient energy in the established formulation.
Here, $M_{\phi}$ is the mobility coefficient, $\Psi$ is the local interface angle, $a(\Psi)$ is the anisotropy coefficient, $H_{\phi}$ is a model constant, $\vartheta$ is an orientation field, and $F_{driv}$ is the driving force.
The model also couples intracellular tubulin transport, competitive tubulin consumption, and neurotrophin dynamics:
\begin{align}
    \frac{\partial (\phi c_{tubu})}{\partial t_{\mathrm{phys}}}
    &= \delta_t \nabla\cdot(\phi \nabla c_{tubu})
    - v_{tub} \cdot \nabla(\phi c_{tubu})
    - k_{tub}(\phi c_{tubu})
    + \epsilon_0
    \frac{|\nabla\phi_0|^2}{\int_{\Omega}|\nabla\phi_0|^2\,d\Omega}, \label{eq:tubulin}\\
    \frac{dL}{dt_{\mathrm{phys}}} &= r_g c_{tubu} - s_g, \label{eq:length}\\
    \frac{\partial c_{neur}}{\partial t_{\mathrm{phys}}}
    &= D_c \nabla^2 c_{neur}
    + (K_{\mathrm{neur}}-k_{p75}c_{neur})\frac{\partial \phi}{\partial t_{\mathrm{phys}}}
    - k_2 c_{neur}. \label{eq:neurotrophin}
\end{align}
The driving force couples these state variables back to the phase-field equation:
\begin{equation}
    F_{driv} = \frac{\alpha_{\mathrm{driv}}}{\pi}
    \tan^{-1}\left[H_{\epsilon}\left(\frac{dL}{dt_{\mathrm{phys}}}\right)\gamma(c_{opti}-c_{neur})\right].
    \label{eq:driving_force}
\end{equation}
Here, $c_{tubu}$ and $c_{neur}$ denote tubulin and neurotrophin concentrations, respectively; $\delta_t$, $v_{tub}$, and $k_{tub}$ control tubulin diffusion, transport velocity, and decay; $r_g$ and $s_g$ are tubulin assembly and disassembly rates; and $D_c$, $K_{\mathrm{neur}}$, $k_{p75}$, and $k_2$ govern neurotrophin transport and reaction.
The normalized source proportional to $|\nabla\phi_0|^2$ localizes tubulin production near the initial interface. In the driving force, $\alpha_{\mathrm{driv}}$ scales its magnitude, $\gamma$ maps the neurotrophin mismatch, and $H_{\epsilon}$ is a regularized switching function.

The numerical discretization follows the locally refined IGA implementation in~\cite{qian2025neurodevelopmental}.
Given a nondecreasing local knot vector $\Xi=\{\xi_i\}$, the univariate B-spline basis is constructed recursively by the Cox--de Boor relations~\cite{piegl2012nurbs}:
\begin{align}
    N_{i,0}(\xi)
    &=
    \begin{cases}
        1, & \xi_i\leq\xi<\xi_{i+1},\\
        0, & \text{otherwise},
    \end{cases} \label{eq:bspline_zero}\\
    N_{i,p}(\xi)
    &=\frac{\xi-\xi_i}{\xi_{i+p}-\xi_i}N_{i,p-1}(\xi)
    +\frac{\xi_{i+p+1}-\xi}{\xi_{i+p+1}-\xi_{i+1}}N_{i+1,p-1}(\xi),
    \label{eq:bspline_recursive}
\end{align}
where a fraction with a zero denominator is taken as zero.
The T-mesh is the locally refinable parametric grid that supports the T-spline basis.
For an anchor $A$ defined on this mesh, the local knot vectors in the two parametric directions determine its tensor-product blending function:
\begin{equation}
    B_A(\xi,\eta)=N_{i,p}(\xi)M_{j,q}(\eta).
    \label{eq:tspline_blending}
\end{equation}
During local refinement, coarse blending functions are truncated by their active refined children:
\begin{equation}
    \widetilde{B}_A
    =B_A-\sum_{B\in\mathcal{R}_A}c_{AB}B'_B,
    \qquad
    \mathcal{N}_A=\frac{\widetilde{B}_A}{\sum_C\widetilde{B}_C},
    \label{eq:truncated_tspline}
\end{equation}
where $\mathcal{R}_A$ contains the refined child functions removed from $B_A$ and $c_{AB}$ are the corresponding refinement coefficients~\cite{wei2017truncated}.
The same normalized basis represents both the computational geometry and each state variable:
\begin{equation}
    \mathbf{x}_h(\boldsymbol{\xi})
    =\sum_{A=1}^{n_b}\mathcal{N}_A(\boldsymbol{\xi})\mathbf{P}_A,
    \qquad
    u_h(\mathbf{x}_h(\boldsymbol{\xi}),t_{\mathrm{phys}})
    =\sum_{A=1}^{n_b}\mathcal{N}_A(\boldsymbol{\xi})u_A(t_{\mathrm{phys}}),
    \label{eq:iga_approximation}
\end{equation}
where $\mathbf{P}_A$ are control points and $u_A(t_{\mathrm{phys}})$ are the time-dependent control variables for $u\in\{\phi,c_{tubu},c_{neur}\}$.
The coupled discrete systems are assembled over the active elements of this locally refined mesh.
Elements intersecting the diffuse phase-field interface are refined while regions where $\phi$ is nearly constant remain coarse; one-ring bisections maintain a strongly balanced mesh and remove inadmissible face--face intersections~\cite{wei2017truncated}.
When a neurite approaches a domain boundary, directional domain expansion adds coarse elements only along the flagged boundary, and a k-d tree transfers the state variables to the updated discretization~\cite{bentley1975multidimensional}.
After each mesh update, the truncated basis and element connectivity are regenerated, and METIS repartitions the resulting irregular computational graph for load balance~\cite{karypis1998fast}. The coupled nonlinear and linear systems are then assembled and solved in parallel with PETSc and MPI~\cite{petsc-user-ref}, using PETSc's scalable communication layer for the required distributed data exchanges~\cite{petscsf2022}.
This adaptive numerical pipeline concentrates degrees of freedom and parallel work near evolving neurite interfaces.
Model parameters and initial and boundary conditions follow the cited IGA study.
The resulting phase-field trajectories provide the synthetic image sequences used for MGRD training and evaluation.

\subsection{Evaluation metrics and statistical analysis}
\label{subsec:evaluation}

All reported predictive-accuracy and sample-variance metrics are computed on the test splits.
For the main comparison, each fitted diffusion model is evaluated with ten independent initial-noise seeds (0--9) using the 50-step DDIM sampler (DDIM-50)~\cite{song2020denoising}.
Each sampled forecast is scored separately over the test split.
For a scalar accuracy metric $g$, the trajectory-wise mean and sample standard deviation (SD) are
\begin{align}
    \overline{g}_{\mathrm{traj}}
    &=\frac{1}{M}\sum_{m=1}^{M}
      g\!\left(\widehat{Y}^{(m)},Y\right),\\
    s_{g,\mathrm{traj}}
    &=\sqrt{\frac{1}{M-1}\sum_{m=1}^{M}
      \left[g\!\left(\widehat{Y}^{(m)},Y\right)
      -\overline{g}_{\mathrm{traj}}\right]^2}.
    \label{eq:trajectory_metric_summary}
\end{align}
These trajectory-wise summaries are used throughout the controlled simulation, experimental, cross-domain transfer, and long-horizon evaluations; gSTA is deterministic and is evaluated once.
All architecture and training controls are evaluated using the same ten initial-noise seeds, and their performance is summarized by $\overline{g}_{\mathrm{traj}}$.
For the controlled simulation comparison, ungated continuation and MGRD begin from the same Stage I checkpoint and use the same Stage-II training seed, optimizer, learning rate, weight decay, batch size, diffusion objective, and five-epoch adaptation budget.
Ungated continuation updates all 243,156 backbone parameters, whereas MGRD freezes the backbone and updates 54,544 morphology-pathway parameters.
These metrics differ from evaluating the pixelwise ensemble mean,
\begin{equation}
    g_{\mathrm{ens}}
    =g\!\left(\overline{Y},Y\right),
    \qquad
    \overline{Y}=\frac{1}{M}\sum_{m=1}^{M}\widehat{Y}^{(m)},
    \label{eq:ensemble_mean_metric}
\end{equation}
because in general $g(\overline{Y},Y)\neq M^{-1}\sum_m g(\widehat{Y}^{(m)},Y)$.
The uncertainty and risk--coverage analyses and ensemble-mean visualization panels use $g_{\mathrm{ens}}$.

All accuracy metrics use the framewise-min--max normalized space after predictions are restricted to $[0,1]$; percentile scaling affects visualization alone.
Each metric is evaluated separately for every predicted frame and then averaged with equal weight over all case--frame pairs in the test split.
Mean absolute error (MAE) is computed by averaging the absolute pixelwise difference within each frame.
Following the prior gSTA study, target-range-normalized root-mean-square error (NRMSE), which was labeled MRE in that work, is defined as
\begin{equation}
    \mathrm{NRMSE} =
    \frac{\sqrt{\frac{1}{HW}\sum_{i,j}(\hat{Y}_{ij}-Y_{ij})^2}}
    {\max(Y)-\min(Y)+\varepsilon_{\mathrm{num}}}\times 100.
\end{equation}
Structural similarity (SSIM) is computed framewise with normalized data range 1.0.
For target-matched morphology evaluation, each target frame supplies an Otsu foreground threshold that is applied to both the target and prediction before skeletonization.
If $S_p$ and $S_t$ denote the resulting prediction and target centerlines, skeleton F1 is computed directly from their overlap as
\begin{equation}
    F_1 = \frac{2\lvert S_p\cap S_t\rvert}
    {\lvert S_p\rvert+\lvert S_t\rvert+\varepsilon_{\mathrm{num}}}.
\end{equation}
For sample-variance analysis, the model generates multiple future sequences for each input~\cite{kendall2017uncertainties}.
Given reconstructed forecast samples $\{\widehat{Y}^{(m)}\}_{m=1}^{M}$ and the ensemble mean in Eq.~\eqref{eq:ensemble_mean_metric}, the pixelwise empirical sample variance is
\begin{equation}
    V=\frac{1}{M-1}\sum_{m=1}^{M}
      \left(\widehat{Y}^{(m)}-\overline{Y}\right)^2.
    \label{eq:pixelwise_sample_variance}
\end{equation}
For case $n$, the scalar case-level variance score and ensemble-mean MAE are
\begin{equation}
    u_n=\frac{1}{T_{\mathrm{out}}HW}\sum_{\nu,i,j}V_{n,\nu ij},
    \qquad
    e_n=\mathrm{MAE}\!\left(\overline{Y}_n,Y_n\right).
    \label{eq:case_variance_error}
\end{equation}
where $\nu$ indexes forecast frames and $i,j$ index spatial pixels.
Thus, $V_n$ is a spatial--temporal variance field, whereas $u_n$ is the scalar ranking score used for selective prediction.
Pearson correlation $r$ measures the linear association between the scalar case-level variance score and ensemble-mean forecast error.
The practical value of variance-based case ordering is evaluated separately through risk--coverage analysis.
Correlation intervals are obtained by 10,000 paired case-level bootstrap resamples and reported as percentile 95\% confidence intervals; both the sample-variance and risk--coverage analyses use $M=10$ independently sampled trajectories per case.

For risk--coverage analysis, cases are ordered from lower to higher scalar variance score and the lowest-variance cases are retained at nominal coverage levels of 100\%, 90\%, 80\%, 70\%, and 60\%.
At each level, the primary risk is the retained-case mean MAE.
The variance ranking is compared with 10,000 random rankings and with an oracle ranking based on the true case-level MAE.
Area under the risk--coverage curve (AURC) is computed by trapezoidal integration over the evaluated coverage range.

Computational efficiency is evaluated through parameter count, sampling latency, and single-step training latency.
Both models are benchmarked on an NVIDIA GeForce RTX 4080 SUPER GPU.
The inference benchmark compares deterministic gSTA with one MGRD DDIM-50 sample using batch size 2 and three repetitions of 20 warmup and 100 timed iterations.
Because the morphology condition is fixed throughout a trajectory, MGRD encodes its two-scale morphology features once and reuses them across all denoising steps; the dynamic gate masks remain timestep-specific.
DDIM-50 is used for all primary comparisons.
A sampler-step study evaluates the fixed final checkpoint at 5, 10, 50, 100, 250, and 500 DDIM steps over the complete 87-case experimental test split using matched initial-noise seeds 0--2; its latency curve uses the same cached-morphology implementation as the final benchmark.
Single-step training latency uses batch size 1, 2 warmup iterations, and 8 timed iterations.

\section{Results and discussion}
\label{sec:results}

\subsection{Controlled simulation evaluation of morphology-gated adaptation}

The phase-field trajectories provide clean, sharply defined interfaces for evaluating whether explicit morphology adaptation improves residual diffusion under controlled geometric conditions.
Table~\ref{tab:simulation_comparison} compares MGRD with ungated continuation under the controlled Stage-II protocol described in Section~\ref{sec:method}.
Across ten DDIM-50 forecasts, MGRD reduces trajectory-wise MAE from $9.96\mathrm{E}{-3}\pm2.39\mathrm{E}{-6}$ under ungated continuation to $9.00\mathrm{E}{-3}\pm2.06\mathrm{E}{-6}$ (a 9.70\% reduction in the mean), while also lowering NRMSE and increasing SSIM and skeleton F1.
This pattern shows that morphology adaptation sharpens local intensity and boundary representation while retaining the global topology already learned in Stage I.
The controlled comparison attributes this improvement to morphology-gated adaptation while reducing the Stage-II update set by 77.6\%.

\begin{table}[!htbp]
\centering
\caption{Controlled phase-field simulation comparison.
Ungated and MGRD entries report the trajectory-wise mean $\pm$ sample SD over ten DDIM-50 forecasts from fixed trained models.
Updated parameters count only those optimized during Stage II.
The relative-change row reports MGRD with respect to ungated continuation.
Arrows indicate the preferred direction; bold denotes the better mean.}
\label{tab:simulation_comparison}
\tableformat
\setlength{\tabcolsep}{6pt}
\renewcommand{\arraystretch}{1.24}
\begin{tabular}{@{}lccccc@{}}
\toprule
Variant & \tablehead{Updated\\params} $\downarrow$ & NRMSE $\downarrow$ & SSIM $\uparrow$ & Skel F1 $\uparrow$ & MAE $\downarrow$ \\
\midrule
\multicolumn{6}{c}{\textit{Residual-diffusion controls}} \\
\midrule
\addlinespace[2pt]
Ungated continuation & 2.43E+5 & \samplestat{3.19E+0}{1.73E-3} & \samplestat{8.05E-1}{1.58E-5} & \samplestat{7.29E-1}{1.72E-4} & \samplestat{9.96E-3}{2.39E-6} \\
MGRD & 5.45E+4 & \samplestat{\tablebest{2.99E+0}}{1.54E-3} & \samplestat{\tablebest{8.41E-1}}{1.72E-5} & \samplestat{\tablebest{7.31E-1}}{1.28E-4} & \samplestat{\tablebest{9.00E-3}}{2.06E-6} \\
Relative change & $-77.6\%$ & $-6.20\%$ & $+4.45\%$ & $+0.25\%$ & $-9.70\%$ \\
\bottomrule
\end{tabular}
\end{table}

\begin{figure}[ht]
\centering
\includegraphics[width=\textwidth]{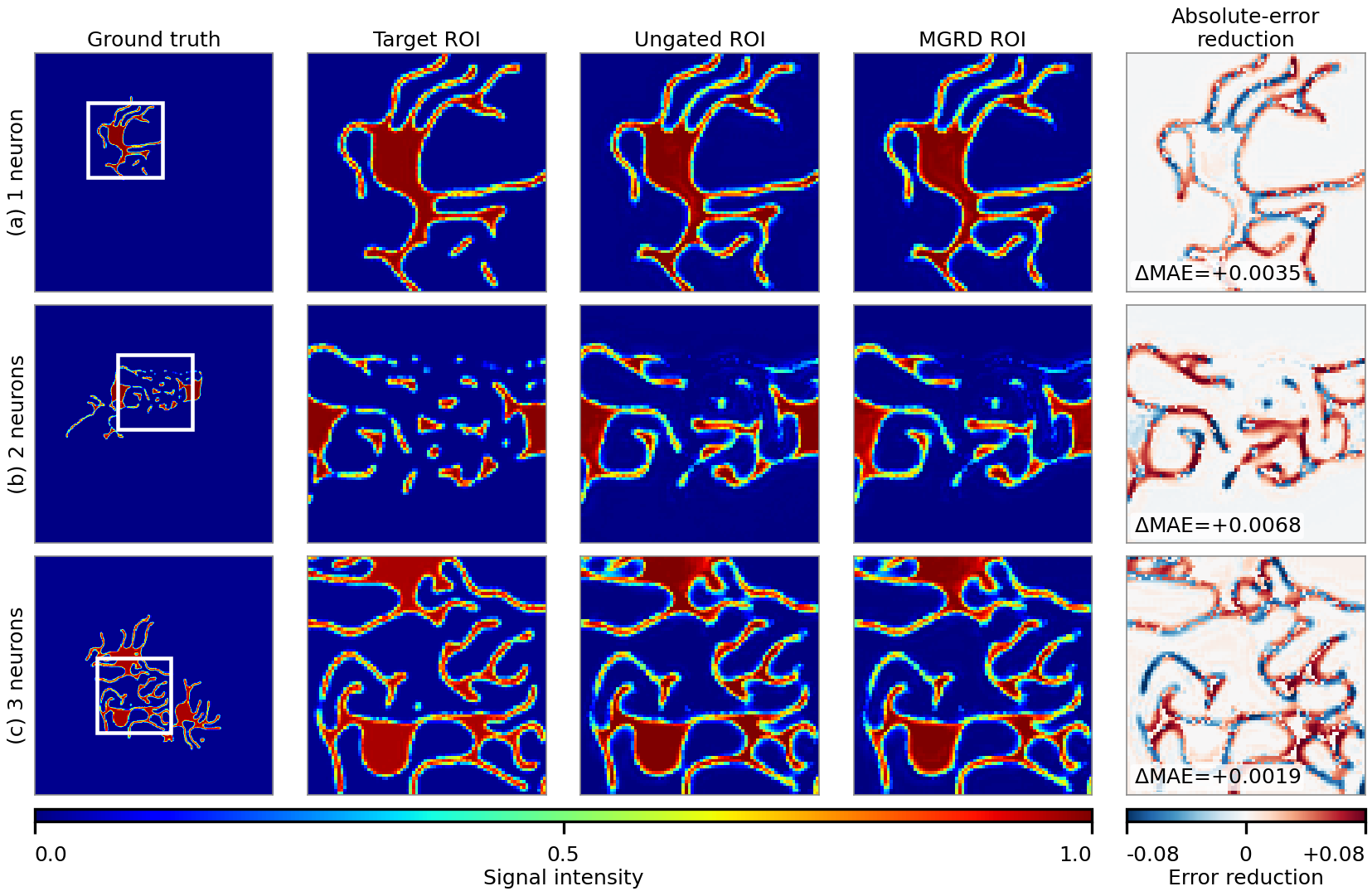}
\caption{Matched five-epoch simulation comparison of morphology-gated local error correction.
Rows (a)--(c) show one-, two-, and three-neuron test trajectories randomly selected from the forecast horizon.
In all three rows, the small disconnected components are simulated neurite remnants arising from the modeled retraction and atrophy process.
In row (b), the ROI is centered midway between the two largest high-intensity soma regions.
The columns show the complete ground truth with a white ROI box, the enlarged target, the pixelwise ensemble means of ten ungated and ten MGRD forecasts, and the paired absolute-error reduction
$\Delta|e|=|Y-\widehat{Y}_{\mathrm{ungated}}|-|Y-\widehat{Y}_{\mathrm{MGRD}}|$.
Positive red values indicate lower absolute error with MGRD, whereas negative blue values indicate lower absolute error with the ungated continuation.
Ground-truth and prediction panels share the fixed 0--1 signal-intensity scale, and all reduction maps share one symmetric scale.
The reported ROI $\Delta$MAE is evaluated within a two-pixel band around the target centerline.}
\label{fig:simulation_multicase_predictions}
\end{figure}

Figure~\ref{fig:simulation_multicase_predictions} localizes the gating comparison across physically interpretable one-, two-, and three-neuron simulation configurations.
Each row begins with the complete target and a target-defined region of interest (ROI), allowing the local comparison to be interpreted within the full morphology.
For the two-neuron example, the ROI covers the processes between the two largest high-intensity soma regions.
The displayed prediction panels are pixelwise ensemble means, and the paired maps show that their differences concentrate along narrow interfaces, curved branches, and junctions rather than the low-signal background.
For the displayed ensemble-mean predictions at the randomly selected frames, MGRD reduces ROI MAE within the two-pixel target-centerline bands by 0.0035, 0.0068, and 0.0019 in the one-, two-, and three-neuron examples, respectively.
The coexistence of positive and negative regions in the reduction maps shows that the aggregate improvements in Table~\ref{tab:simulation_comparison} arise from spatially and case-dependent corrections rather than uniform gains.

\subsection{Experimental microscopy predictive performance}

\begin{figure}[p]
\centering
\captionsetup{skip=4pt}
\includegraphics[width=0.92\textwidth]{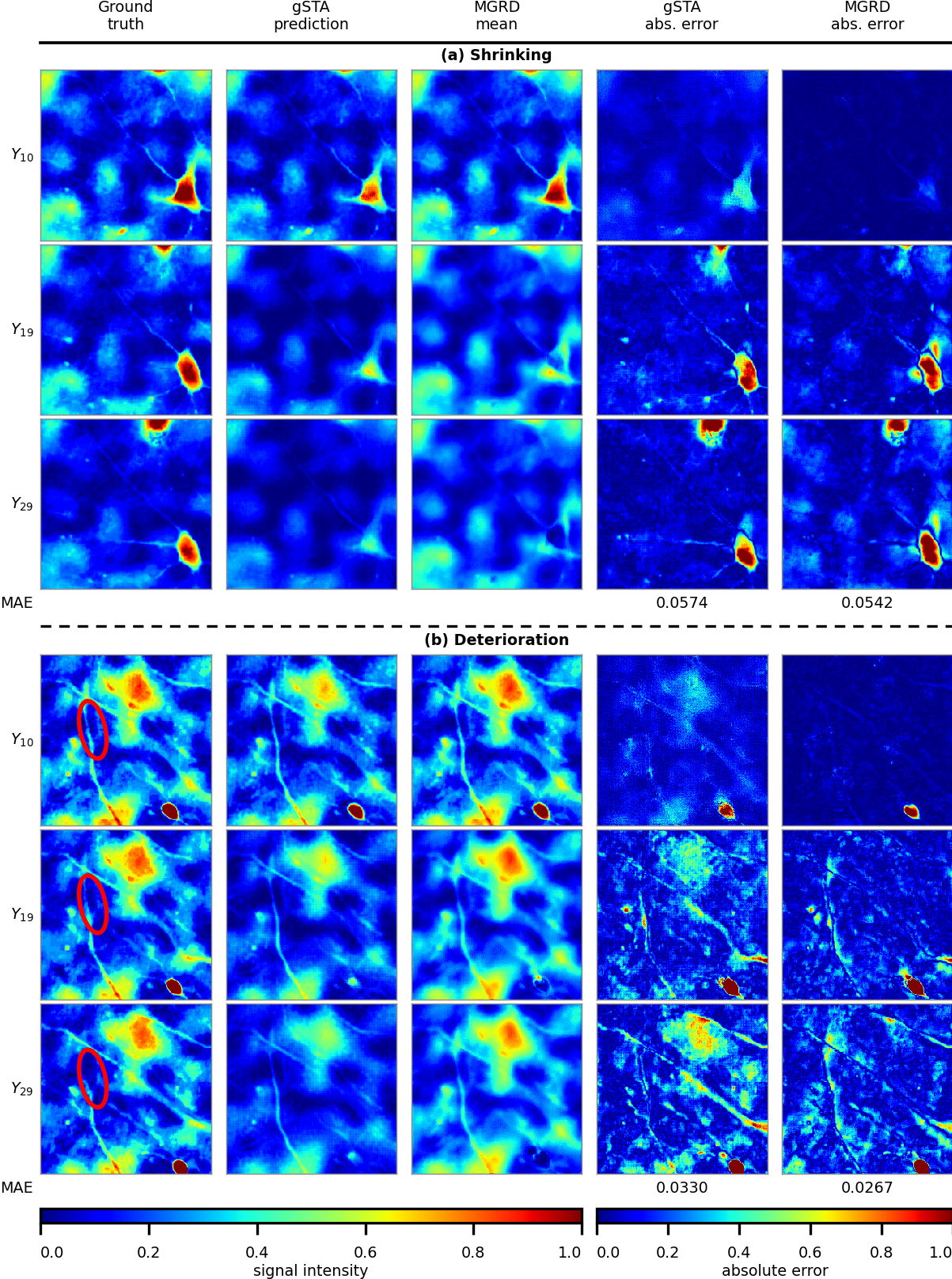}
\caption{Temporal forecasts for (a) a shrinking case and (b) a deterioration case.
Rows show $Y_{10}$, $Y_{19}$, and $Y_{29}$ after the observed history $X_0,\ldots,X_9$.
Columns show ground truth, gSTA prediction, the pixelwise mean of ten MGRD forecasts, and their absolute errors.
The red ellipse marks the connection-loss region; MAE summarizes the full 20-frame forecast for each case.
Within each case, signal and error scales are fixed separately.}
\label{fig:experimental_multicase_predictions}
\end{figure}

Figure~\ref{fig:experimental_multicase_predictions} compares two held-out experimental trajectories across three fixed forecast horizons. After the observed history $X_0,\ldots,X_9$, the panels show the first, tenth, and twentieth jointly predicted target frames $Y_{10}$, $Y_{19}$, and $Y_{29}$. The shrinking sequence, retained from the gSTA study~\cite{qian2025high}, exhibits progressive contraction and weakening of the high-intensity cellular region. In the deterioration sequence, the rotated red ellipse follows the slender process descending from the upper-left junction. Its continuity and contrast progressively weaken over the forecast. MGRD yields less extensive absolute-error regions throughout this sequence, reducing the 20-frame MAE from 0.0330 for gSTA to 0.0267. The qualitative examples localize these differences, whereas the aggregate comparison in Table~\ref{tab:main_comparison} is evaluated over the complete test split.

\begin{table}[!htbp]
\centering
\caption{Experimental microscopy predictive performance of gSTA and MGRD.
MGRD entries summarize ten DDIM-50 forecasts using the convention defined in Section~\ref{subsec:evaluation}.
gSTA is deterministic and reported once.
Arrows indicate the preferred direction; bold denotes the better mean.}
\label{tab:main_comparison}
\tableformat
\setlength{\tabcolsep}{7pt}
\renewcommand{\arraystretch}{1.18}
\begin{tabular}{lcccc}
\toprule
Method & NRMSE $\downarrow$ & SSIM $\uparrow$ & Skel F1 $\uparrow$ & MAE $\downarrow$ \\
\midrule
\addlinespace[2pt]
gSTA & 8.72E+0 & 6.50E-1 & 1.55E-1 & 6.41E-2 \\
MGRD & \samplestat{\tablebest{6.12E+0}}{2.29E-2} & \samplestat{\tablebest{7.42E-1}}{2.58E-4} & \samplestat{\tablebest{2.26E-1}}{1.41E-3} & \samplestat{\tablebest{3.87E-2}}{5.99E-5} \\
\bottomrule
\end{tabular}
\end{table}

Table~\ref{tab:main_comparison} shows that MGRD improves all four experimental metrics under the trajectory-wise protocol: relative to deterministic gSTA ($6.41\mathrm{E}{-2}$), MGRD's MAE is $3.87\mathrm{E}{-2}\pm5.99\mathrm{E}{-5}$, a 39.6\% reduction in the mean.
The NRMSE score also decreases, while SSIM and skeleton F1 increase.
The simultaneous gains show that MGRD improves centerline preservation alongside image fidelity.
The lower MAE and NRMSE quantify the reduced intensity discrepancy visible in the error maps, the SSIM gain reflects better preservation of local image organization, and the largest relative improvement in skeleton F1 is consistent with the explicit centerline-based conditioning pathway.

Table~\ref{tab:mgrd_ablation} further separates the contributions of optimization, multi-scale gating, initialization, and parameter freezing on the experimental data.
All continuation and adapter variants begin from the same Stage I backbone state, whereas the gated-from-scratch control is trained under a 250-epoch schedule.
Continuation of the ungated backbone substantially reduces MAE relative to the Stage-I model alone, showing that Stage-II adaptation is an important source of improvement.
The continuation variants outperform gated training from scratch, supporting reuse of the residual-diffusion representation learned in Stage I.
Among the adapter-only variants, zero-initialized MGRD attains lower MAE and NRMSE and higher SSIM and skeleton F1 than random initialization while updating the same 54,544 parameters.
Zero initialization preserves the pretrained backbone mapping at the start of Stage II and allows the morphology pathway to enter as a learned residual correction.
Together, these controls support the combined adaptation design of pretrained-backbone reuse, zero-initialized multi-scale gates, and adapter-only optimization.

\begin{table}[!htbp]
\centering
\caption{Experimental architecture and training controls under the common evaluation protocol.
Updated parameters count the parameters optimized under the training protocol listed for each variant.
All stochastic variants use the same ten DDIM-50 initial-noise seeds; bold denotes the best mean.}
\label{tab:mgrd_ablation}
\tableformat
\renewcommand{\arraystretch}{1.18}
\begin{tabular}{@{}>{\scriptsize}l@{\hspace{5pt}}>{\scriptsize}l@{\hspace{5pt}}>{\scriptsize}c@{\hspace{14pt}}c@{\hspace{16pt}}c@{\hspace{16pt}}c@{\hspace{16pt}}c@{}}
\toprule
Variant & Optimization protocol & Updated params & MAE $\downarrow$ & NRMSE $\downarrow$ & SSIM $\uparrow$ & Skel F1 $\uparrow$ \\
\midrule
\addlinespace[2pt]
\meanlabel{Ungated backbone} & \meanlabel{Stage I only} & \meanlabel{2.43E+5} & \stackstat[t]{6.13E-2}{1.75E-4} & \stackstat[t]{8.33E+0}{3.13E-2} & \stackstat[t]{6.83E-1}{5.70E-4} & \stackstat[t]{2.03E-1}{8.00E-4} \\
\meanlabel{Ungated continuation} & \meanlabel{Full-model continuation} & \meanlabel{2.43E+5} & \stackstat[t]{3.94E-2}{7.80E-5} & \stackstat[t]{6.19E+0}{2.61E-2} & \stackstat[t]{7.37E-1}{3.16E-4} & \stackstat[t]{2.25E-1}{1.19E-3} \\
\meanlabel{Gated from scratch} & \meanlabel{End-to-end, 250 epochs} & \meanlabel{2.98E+5} & \stackstat[t]{4.41E-2}{8.18E-6} & \stackstat[t]{6.48E+0}{9.70E-4} & \stackstat[t]{6.87E-1}{7.33E-5} & \stackstat[t]{1.84E-1}{2.82E-4} \\
\meanlabel{Joint gated optimization} & \meanlabel{Full-model adaptation} & \meanlabel{2.98E+5} & \stackstat[t]{3.94E-2}{7.80E-5} & \stackstat[t]{6.19E+0}{2.61E-2} & \stackstat[t]{7.37E-1}{3.16E-4} & \stackstat[t]{2.25E-1}{1.21E-3} \\
\meanlabel{Randomly initialized adapter} & \meanlabel{Adapter-only adaptation} & \meanlabel{5.45E+4} & \stackstat[t]{3.88E-2}{5.59E-5} & \stackstat[t]{6.20E+0}{2.28E-2} & \stackstat[t]{7.39E-1}{2.52E-4} & \stackstat[t]{2.20E-1}{1.40E-3} \\
\meanlabel{MGRD} & \meanlabel{Adapter-only adaptation} & \meanlabel{5.45E+4} & \stackstat[t]{\tablebest{3.87E-2}}{5.99E-5} & \stackstat[t]{\tablebest{6.12E+0}}{2.29E-2} & \stackstat[t]{\tablebest{7.42E-1}}{2.58E-4} & \stackstat[t]{\tablebest{2.26E-1}}{1.41E-3} \\
\bottomrule
\end{tabular}
\end{table}

\subsection{Cross-dataset transfer and long-horizon robustness}
\label{subsec:cross_domain}

\begin{figure}[!htbp]
\centering
\includegraphics[width=0.65\textwidth]{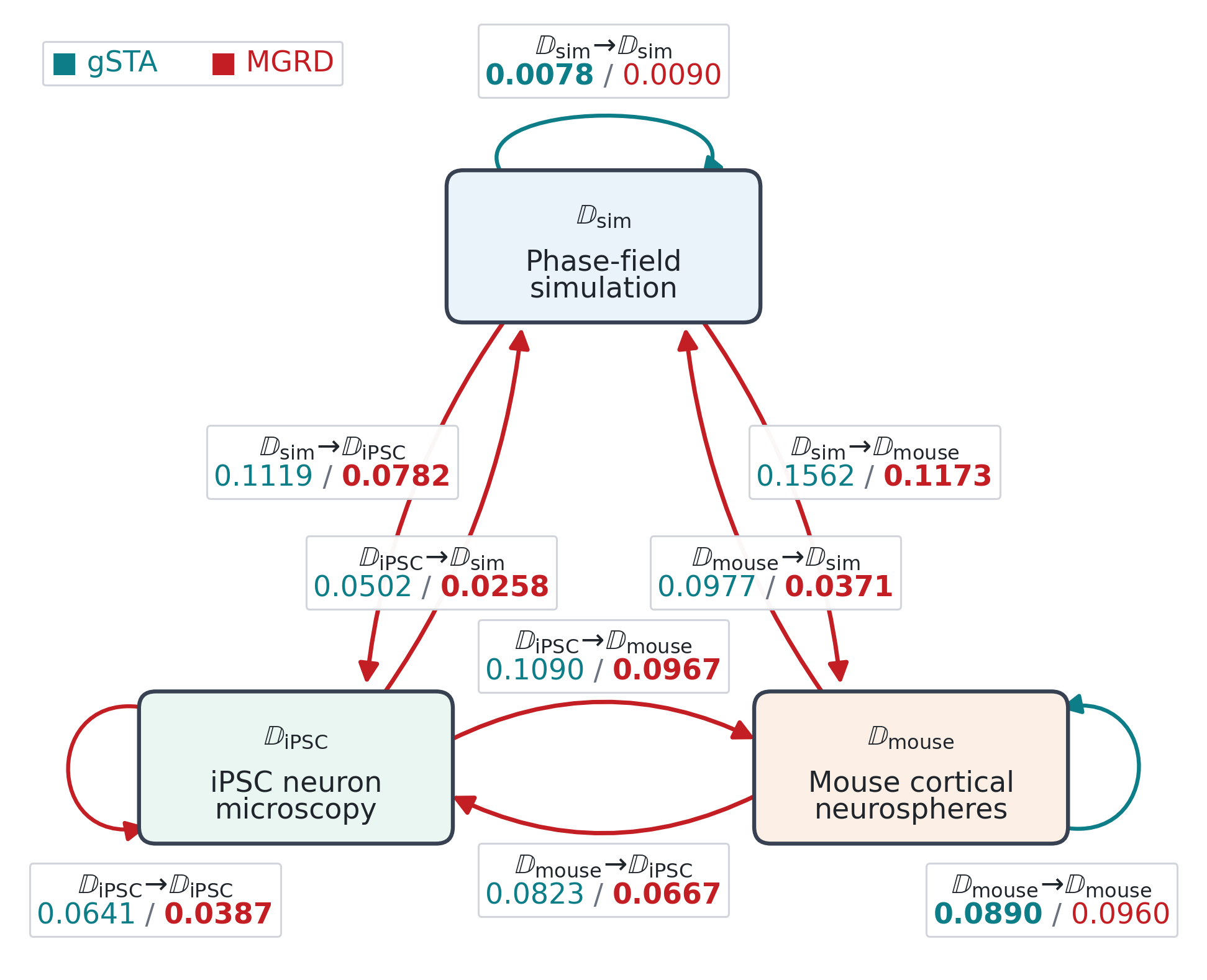}
\caption{Directional transfer among the phase-field simulation ($\Dsim$), iPSC-derived neuron microscopy ($\Dipsc$), and mouse cortical-neurosphere microscopy ($\Dmouse$) datasets at the native 10-min cadence.
Arrows point from the training domain to the inference domain, and self-loops denote in-domain evaluation.
Each arrow label gives its direction on the first line, followed by deterministic gSTA MAE in teal and trajectory-wise mean MAE across ten MGRD DDIM-50 forecasts in red.
The lower MAE is bold, and each arrow uses the color of the better-performing model for that direction.}
\label{fig:cross_dataset_transfer}
\end{figure}

Cross-dataset transfer asks whether a model trained on one data source can forecast sequences from another without retraining or target-domain fine-tuning.
For concise notation in this section, the phase-field simulation, iPSC-derived neuron microscopy, and mouse cortical-neurosphere microscopy datasets are denoted $\Dsim$, $\Dipsc$, and $\Dmouse$, respectively.
It is a stricter test than within-dataset evaluation because these datasets differ in both image appearance and growth dynamics.
We therefore trained models in each domain and evaluated all nine training--inference combinations at the native 10-min cadence, as summarized in Figure~\ref{fig:cross_dataset_transfer}.
Self-loops show performance on each source domain's held-out test split, while the arrows between domains evaluate the corresponding source-trained model directly on another domain's test split.
This design separates transfer direction from target difficulty because $\Dsim\!\rightarrow\!\Dmouse$ and $\Dipsc\!\rightarrow\!\Dmouse$ share the same $\Dmouse$ test cases, while $\Dsim\!\rightarrow\!\Dipsc$ and $\Dmouse\!\rightarrow\!\Dipsc$ share the same $\Dipsc$ test cases.

The directional results reveal a consistent effect of similarity between the training and evaluation domains.
For both experimental targets, a model trained on the other microscopy dataset is more accurate than a model trained only on phase-field simulations.
Across the two models and both experimental targets, replacing $\Dsim$ training with training on the other microscopy dataset reduces MAE by between 14.6\% and 30.2\%.
Thus, under this protocol, information learned from one experimental imaging domain transfers more effectively to another experimental domain than information learned from the synthetic trajectories alone.

The comparison between models depends on whether the evaluation remains within the training domain or crosses into a new one.
gSTA has lower MAE for the $\Dsim\!\rightarrow\!\Dsim$ and $\Dmouse\!\rightarrow\!\Dmouse$ evaluations, whereas MGRD has lower MAE for $\Dipsc\!\rightarrow\!\Dipsc$ and all six transfers between different datasets.
The advantage of MGRD is therefore most consistent when the training and evaluation distributions differ.
To examine the $\Dipsc\!\rightarrow\!\Dmouse$ transfer over longer horizons, we evaluated the iPSC-trained models at 10-, 20-, and 40-min sampling intervals without mouse-domain fine-tuning in Table~\ref{tab:external_stride}.
At the 40-min interval, where the final target lies 13 h 20 min beyond the latest observation, MGRD's trajectory-wise mean MAE across the full forecast sequence was 11.2\% lower than gSTA's on the evaluated test set.
\begin{table}[!htbp]
\centering
\caption{Cross-domain evaluation on mouse cortical-neurosphere microscopy at increasing temporal sampling intervals.
MGRD entries summarize ten DDIM-50 forecasts using the convention defined in Section~\ref{subsec:evaluation}; gSTA is deterministic and evaluated once.
The 10-min row uses the fixed specimen-disjoint mouse test split, whereas the 20- and 40-min rows use the available long-horizon cases from the same mouse collection.
Bold indicates the best value within each sampling-interval block.}
\label{tab:external_stride}
\tableformat
\setlength{\tabcolsep}{5.5pt}
\begin{tabular}{llrrrr}
\toprule
Interval (min) & Method & MAE $\downarrow$ & NRMSE $\downarrow$ & SSIM $\uparrow$ & Skel F1 $\uparrow$ \\
\midrule
\multirow{2}{*}{\raisebox{-5.3pt}[0pt][0pt]{10}} & gSTA & 1.09E-1 & 1.46E+1 & 1.85E-1 & 1.98E-2 \\
\addlinespace[3pt]
 & \shortstack[t]{\raisebox{1.3pt}{MGRD}\\\phantom{$\pm$}} & \stackstat[t]{\tablebest{9.67E-2}}{9.68E-5} & \stackstat[t]{\tablebest{1.26E+1}}{1.40E-2} & \stackstat[t]{\tablebest{2.68E-1}}{1.61E-4} & \stackstat[t]{\tablebest{1.82E-1}}{1.99E-4} \\
\midrule
\multirow{2}{*}{\raisebox{-5.3pt}[0pt][0pt]{20}} & gSTA & 5.32E-2 & \tablebest{7.33E+0} & 4.14E-1 & 1.23E-1 \\
\addlinespace[3pt]
 & \shortstack[t]{\raisebox{1.3pt}{MGRD}\\\phantom{$\pm$}} & \stackstat[t]{\tablebest{4.95E-2}}{1.08E-4} & \stackstat[t]{7.40E+0}{2.96E-2} & \stackstat[t]{\tablebest{4.60E-1}}{2.54E-4} & \stackstat[t]{\tablebest{1.63E-1}}{3.65E-4} \\
\midrule
\multirow{2}{*}{\raisebox{-5.3pt}[0pt][0pt]{40}} & gSTA & 5.61E-2 & 7.65E+0 & 3.78E-1 & 9.15E-2 \\
\addlinespace[3pt]
 & \shortstack[t]{\raisebox{1.3pt}{MGRD}\\\phantom{$\pm$}} & \stackstat[t]{\tablebest{4.98E-2}}{1.48E-4} & \stackstat[t]{\tablebest{7.59E+0}}{3.01E-2} & \stackstat[t]{\tablebest{4.32E-1}}{2.20E-4} & \stackstat[t]{\tablebest{1.17E-1}}{3.02E-4} \\
\bottomrule
\end{tabular}
\end{table}

\begin{figure}[p]
\centering
\includegraphics[width=0.98\textwidth]{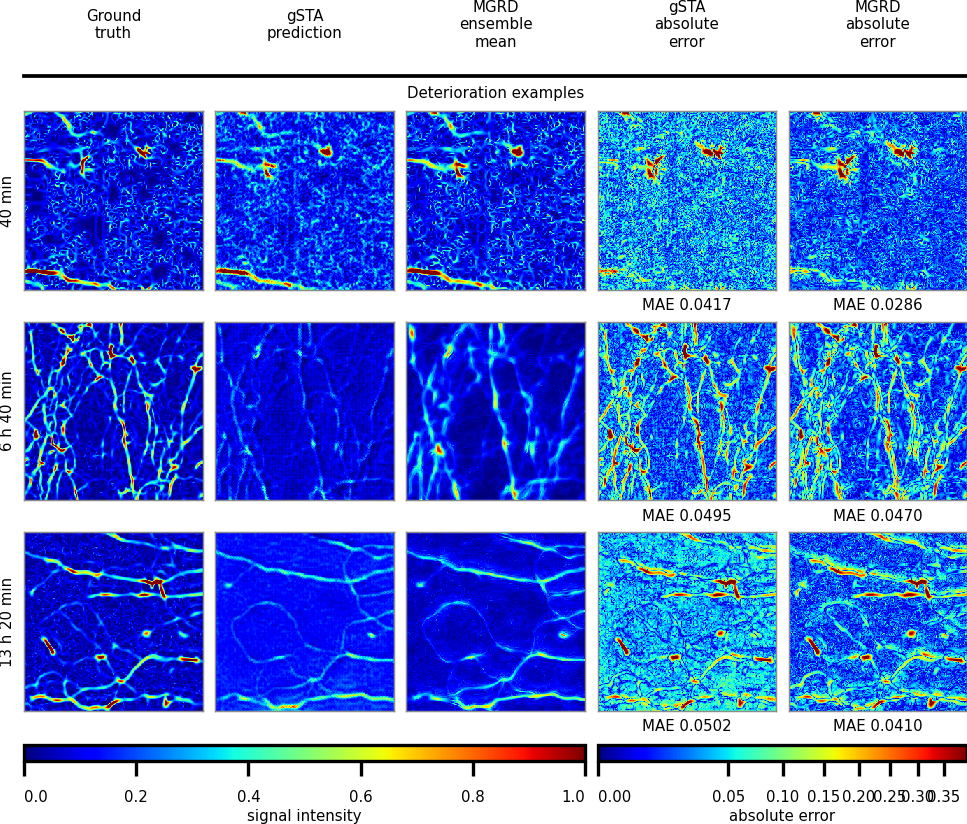}
\caption{Illustrative deterioration forecasts for mouse cortical-neurosphere microscopy at a 40-min sampling interval.
Rows show three different spatial cases at forecast lead times of 40 min, 6 h 40 min, and 13 h 20 min from the latest input; the cases were selected such that the displayed frame at each horizon has lower MAE for the MGRD ensemble mean than for gSTA.
Columns show ground truth, deterministic gSTA prediction, the pixelwise ensemble mean of ten MGRD forecasts, and the corresponding absolute-error maps.
MAE is reported for each displayed frame.
All signal panels share one display scale, and both error columns share another.
The error maps use a common square-root contrast transform to reveal low-amplitude discrepancies.}
\label{fig:external_long_horizon}
\end{figure}

The aggregate long-horizon advantage in Table~\ref{tab:external_stride} is examined spatially in Figure~\ref{fig:external_long_horizon} using three illustrative deterioration cases from the 40-min setting, one at each displayed forecast horizon.
The cases were selected such that the displayed frame at each horizon has lower MAE for the MGRD ensemble mean than for gSTA; they illustrate the spatial character of the improvement rather than replace the aggregate comparison in Table~\ref{tab:external_stride}.
At the final 13 h 20 min horizon, the MGRD ensemble mean more accurately retains the attenuating neurite morphology and reduces the displayed-frame MAE from $5.02\mathrm{E}{-2}$ to $4.10\mathrm{E}{-2}$.
Across the displayed cases, the MGRD error is weaker in low-signal regions and more concentrated around the observed neurite structure than the broader gSTA error.
This behavior is consistent with the skeleton- and distance-conditioned residual denoiser suppressing changes that lack support from the observed morphology.

\begin{figure}[p]
\centering
\includegraphics[width=0.98\textwidth]{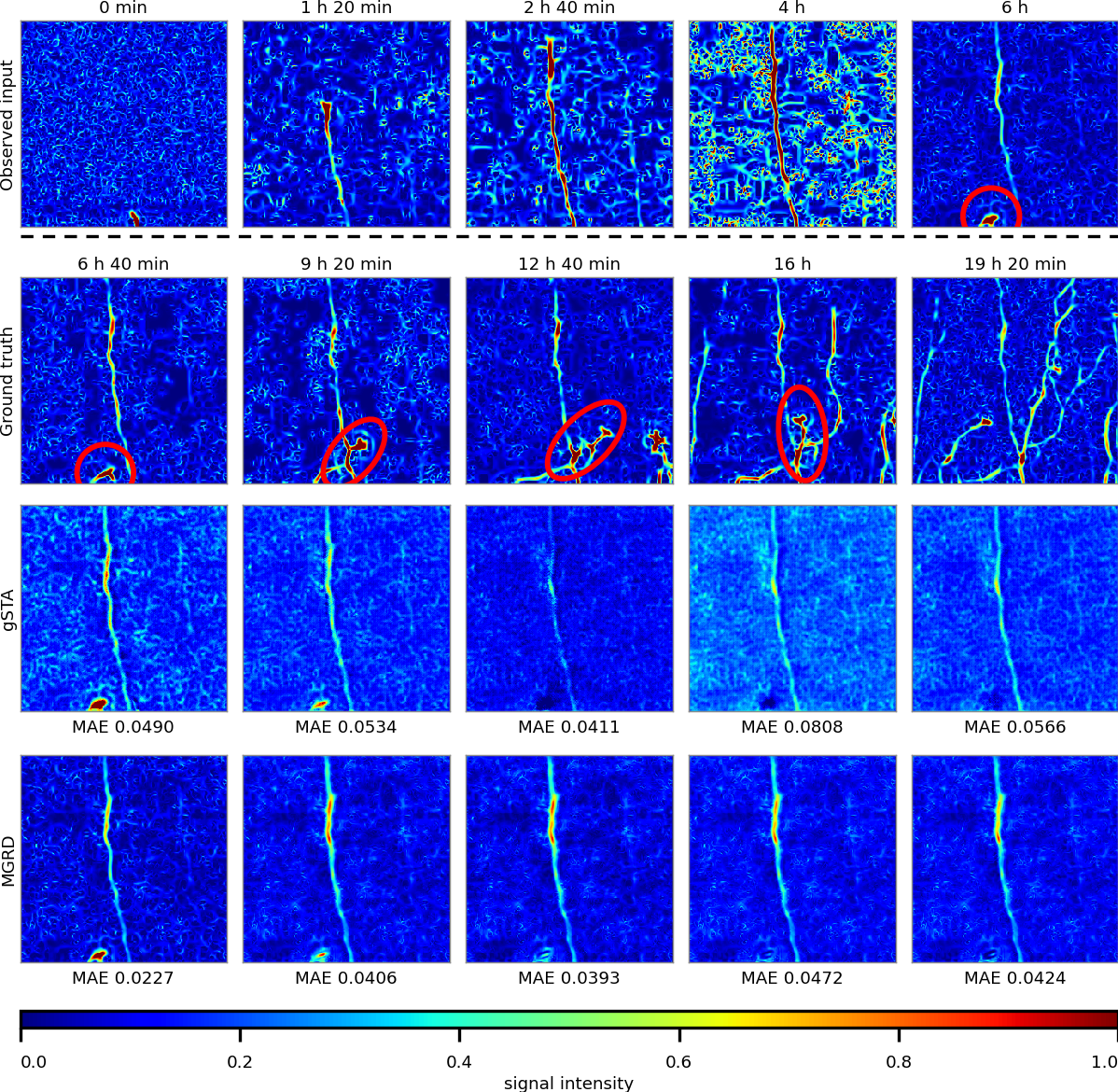}
\caption{Incoming-neurite failure case in mouse cortical-neurosphere microscopy at a 40-min sampling interval.
The first row shows five selected frames from the ten-frame observed history, and the second row shows five selected frames from the twenty-frame future ground-truth sequence; times are measured from the first observed frame.
Red outlines identify a neurite entering through the lower crop boundary and its subsequent extension.
The final two rows show deterministic gSTA predictions and the pixelwise ensemble mean of ten MGRD forecasts, with MAE reported for each displayed target frame.
All signal panels share the display scale derived from the observed inputs.}
\label{fig:incoming_neurite_failure}
\end{figure}

A second case illustrates a distinct failure mode: neurites that enter the crop from outside the observed field of view.
In Figure~\ref{fig:incoming_neurite_failure}, the red outline marks a neurite tip that becomes marginally visible at the lower boundary only in the final observed frame and then extends substantially into the crop during the target sequence.
Because the future branch geometry is absent from the preceding input history, neither model reconstructs its subsequent extension accurately; both instead favor continuation of the neurite scaffold already present in the observed frames.
The gSTA forecasts additionally exhibit a broad background shift, while MGRD maintains a cleaner continuation of the pre-existing scaffold.
This case illustrates a limitation of crop-based forecasting: a neurite tip visible only at the image boundary provides limited information about its subsequent extension into the crop.

\subsection{Sample variance and selective forecasting}
\label{subsec:uncertainty_results}

Repeated MGRD sampling provides an empirical disagreement score for prioritizing forecasts that may require closer review.
For this analysis, forecast MAE is computed from the pixelwise ensemble mean of ten DDIM-50 trajectories, and cases are ranked by the scalar case-level variance score in Eq.~\eqref{eq:case_variance_error}.
At approximately 60\% coverage, retaining the cases with the lowest variance reduces MAE by 17.6\% on the experimental data and 16.8\% on the simulation data relative to retaining all cases.
The same ordering reduces MAE AURC relative to random ranking by 8.4\% and 7.1\%, respectively.
Both comparisons show that low variance identifies subsets with more accurate ensemble forecasts across the two model development domains.

\begin{figure}[!htbp]
\centering
\includegraphics[width=\textwidth]{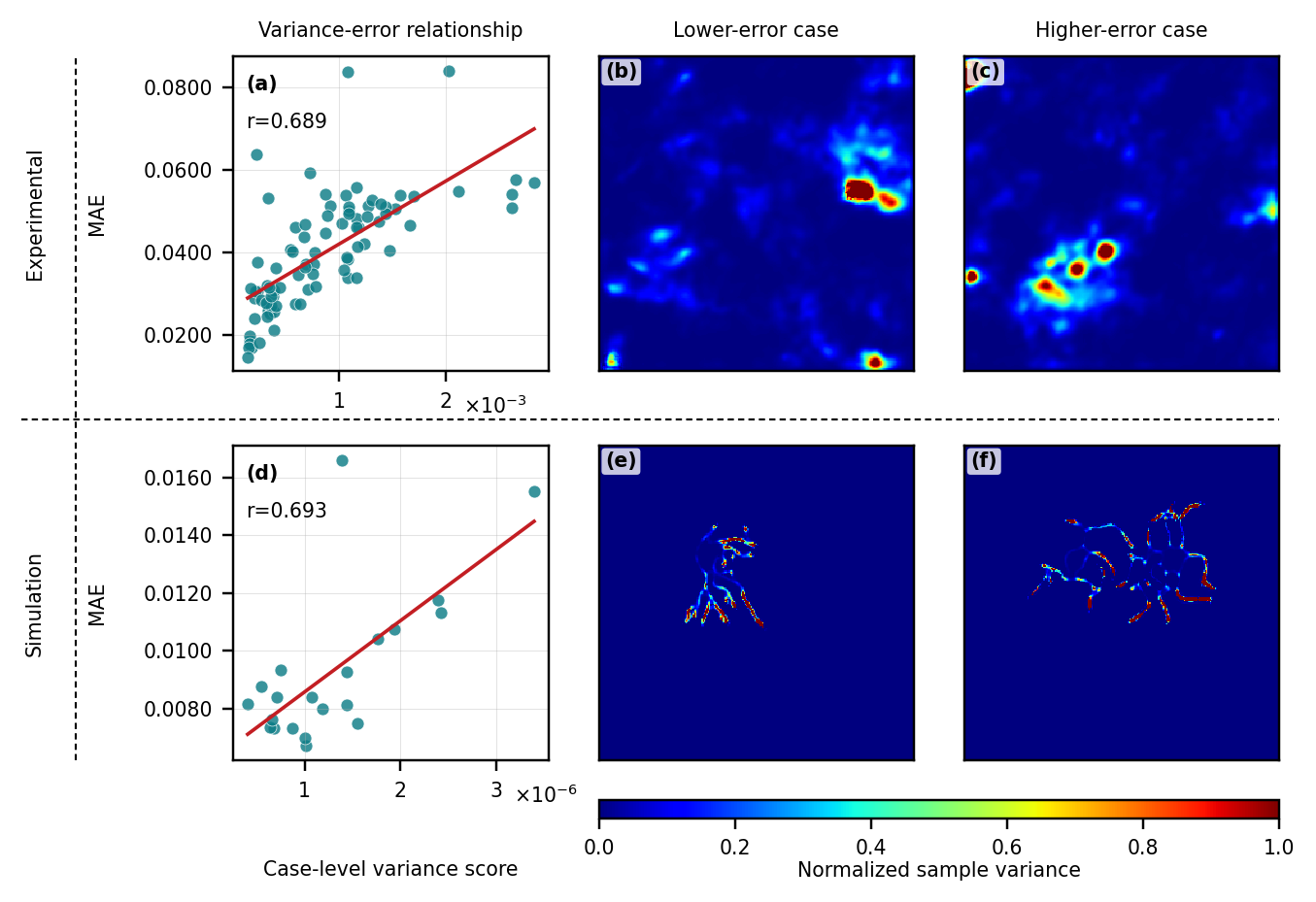}
\caption{Case-level variance--error relationship for MGRD using ten DDIM-50 samples.
(a) Experimental scalar case-level variance score versus ensemble-mean MAE; (b) and (c) final-frame spatial variance maps for representative lower- and higher-error experimental cases.
(d) Simulation scalar case-level variance score versus ensemble-mean MAE; (e) and (f) final-frame spatial variance maps for representative lower- and higher-error simulation cases.
Panels (a) and (d) report Pearson correlation.
Maps are normalized within each dataset to a common row-wise scale for visualization; correlations and errors are computed from unscaled values.}
\label{fig:uncertainty_error}
\end{figure}

Figure~\ref{fig:uncertainty_error} further shows that the scalar case-level variance score is positively associated with ensemble-mean MAE on both experimental data (Pearson $r=0.645$, 95\% confidence interval (CI) $[0.522,0.769]$) and simulation data (Pearson $r=0.664$, 95\% CI $[0.411,0.950]$).
The spatial variance maps help explain these case-level associations by showing where the sampled trajectories disagree.
In the representative experimental maps, substantial variance occurs near the crop boundaries in both the lower- and higher-error cases (Figure~\ref{fig:uncertainty_error}(b,c)), although it is more spatially extensive in the higher-error case.
This boundary localization is consistent with a limitation of crop-based forecasting: neurons or neurites that lie partly outside the observed field of view can enter the crop during the forecast period, making their future appearance weakly constrained by the input history and increasing disagreement among sampled trajectories.
In the simulation maps (Figure~\ref{fig:uncertainty_error}(e,f)), the strongest variance is concentrated around neurite tips and thin terminal interfaces in both error regimes.
These are the locations at which small differences in the predicted extent or direction of retraction produce the largest spatial disagreement, while the persistent interior morphology remains comparatively stable.

\begin{figure}[!htbp]
\centering
\includegraphics[width=\textwidth]{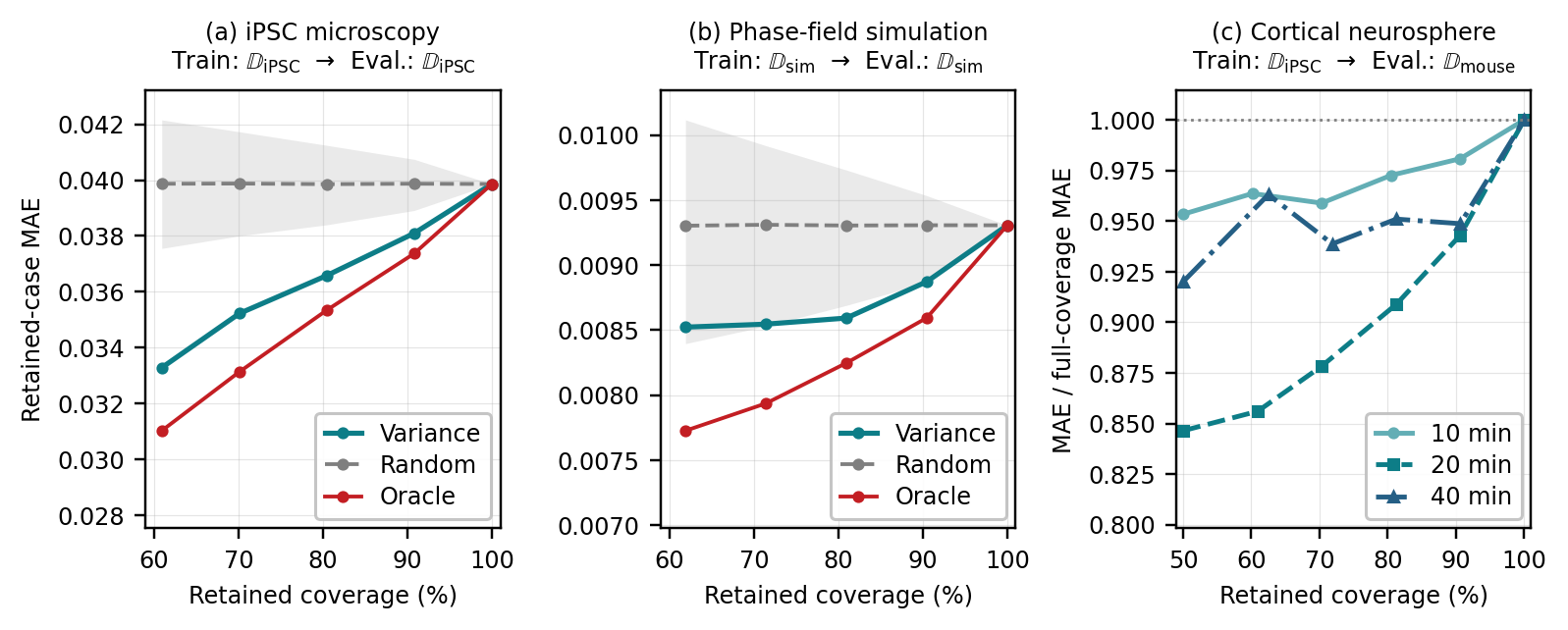}
\caption{Risk--coverage analysis using the scalar MGRD case-level variance score.
Panels (a) and (b) show the in-domain $\Dipsc\!\rightarrow\!\Dipsc$ and $\Dsim\!\rightarrow\!\Dsim$ evaluations, respectively: teal curves retain the lowest-variance cases, gray curves and bands show the mean and pointwise 95\% range from 10,000 random rankings, and red curves show the MAE-based oracle ranking.
Panel (c) shows $\Dipsc\!\rightarrow\!\Dmouse$ evaluation without target-domain fine-tuning and compares variance-ranked results at 10-, 20-, and 40-min sampling intervals; retained MAE is normalized by the full-coverage MAE within each interval.}
\label{fig:risk_coverage}
\end{figure}

Figure~\ref{fig:risk_coverage} extends the selective prediction analysis to cortical-neurosphere microscopy at sampling intervals of 10, 20, and 40 min.
Variance ranking improves on random ranking at every interval, reducing MAE AURC by between 3.0\% and 10.0\%.
The clearest separation occurs at 20 min, where retaining approximately 60\% of the cases reduces MAE by 14.4\% relative to full coverage and the association between variance and error reaches Pearson $r=0.606$ (95\% CI $[0.416,0.752]$).
The improvements are smaller at 10 and 40 min, but their direction remains consistent.
This result extends the association observed in the two model development domains to transferred forecasts and shows that cases can be prioritized without using the unknown target at selection time.
Across these experiments, sample variance is therefore best interpreted as a relative ranking signal that helps identify more reliable forecasts, rather than as a calibrated estimate of error.

\subsection{DDIM sampling budget and computational efficiency}

\begin{figure}[!htbp]
\centering
\includegraphics[width=\textwidth]{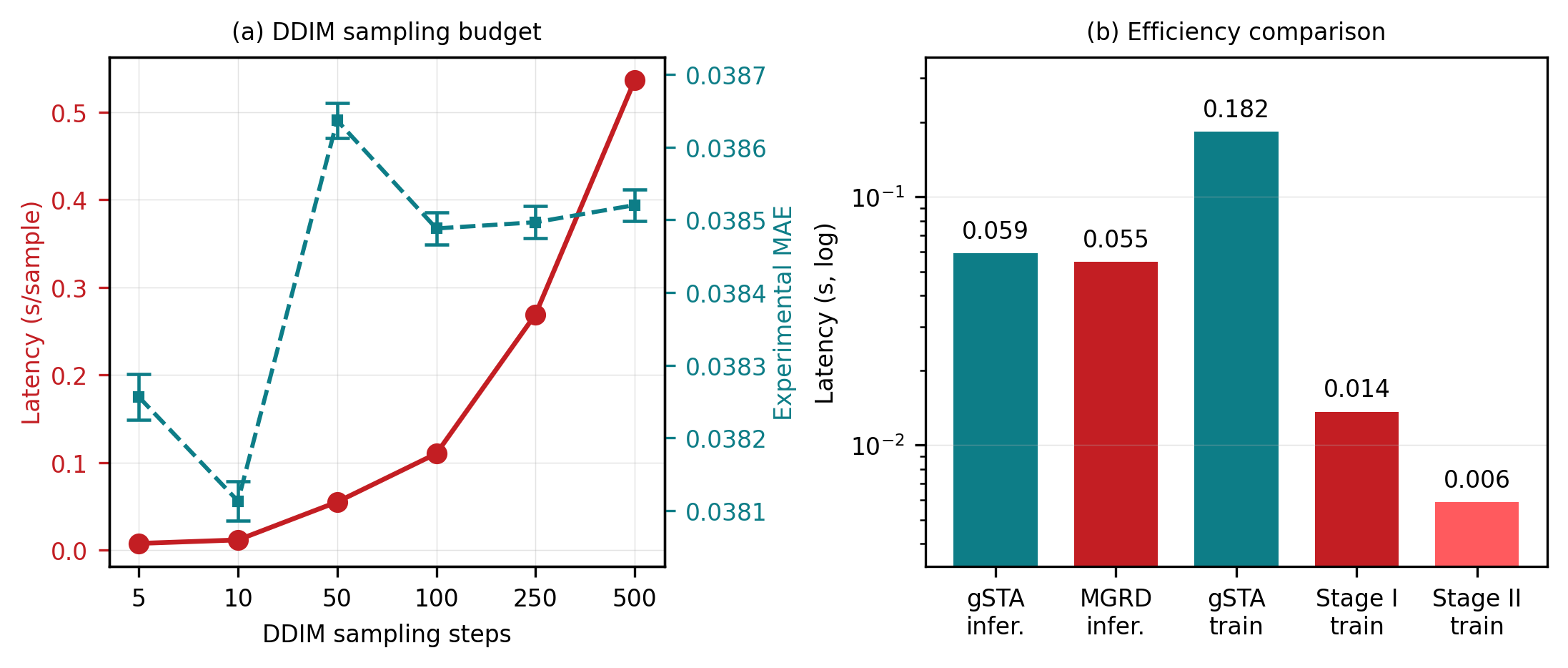}
\caption{Computational efficiency summary.
(a) Effect of the DDIM sampling budget on cached latency for one trajectory and mean experimental test MAE for the fixed final MGRD checkpoint; MAE error bars span the mean $\pm$ the sample SD across three initial-noise seeds.
(b) Measured inference latency for one trajectory and latency for one training update; only the morphology adapter parameters are optimized in Stage II.}
\label{fig:efficiency_summary}
\end{figure}

Figure~\ref{fig:efficiency_summary}(a) examines how the DDIM sampling budget affects experimental forecast accuracy and computation time.
The results show that increasing the number of denoising steps has little influence on accuracy but a substantial effect on sampling cost.
Specifically, as the budget increases from 5 to 500 steps, the mean experimental MAE across three matched initial-noise seeds varies only from $3.81\mathrm{E}{-2}$ to $3.86\mathrm{E}{-2}$, with a sample SD of $2.15\mathrm{E}{-5}$ to $3.17\mathrm{E}{-5}$ across the tested step counts, and exhibits no consistent downward trend.
Over the same range, the time required to generate each sample increases by a factor of 70.6.
The limited variation in MAE, together with the absence of a monotonic improvement, indicates that the forecasts have already reached a stable accuracy regime at relatively short schedules for this checkpoint.
Consequently, the setting of 50 DDIM steps used in the main experiments provides accuracy comparable to that of the longer schedules while avoiding their substantially greater computational cost.

Figure~\ref{fig:efficiency_summary}(b) places this sampling cost in the context of model training and gSTA inference.
MGRD contains 297,700 parameters, which is 1.01\% of the 29.4 million parameters in gSTA, and Stage II optimizes only 54,544 parameters in the morphology pathway.
On the common RTX 4080 SUPER benchmark, the time required for one Stage I or Stage II update is lower than for a gSTA update by factors of 13.4 and 31.0, respectively.
In the three-repeat inference benchmark, one cached MGRD trajectory with 50 DDIM steps takes $54.57\pm0.49$~ms, compared with $59.26\pm0.82$~ms for one deterministic gSTA forecast, a 7.9\% reduction in mean latency.
An ensemble repeats this operation, but the resulting trajectories also provide the sample variance used to rank cases in Section~\ref{subsec:uncertainty_results}, which is not available from the deterministic baseline.
Taken together, these measurements show that MGRD can produce stochastic forecasts and estimate predictive uncertainty without relying on a larger model or slower individual forecasts.
Its compact architecture reduces the cost of model fitting and maintains the speed of each forecast.
The remaining computational cost is controlled by the chosen ensemble size, with additional samples used specifically to estimate uncertainty.

\section{Conclusions and future work}
\label{sec:conclusion}

This study establishes MGRD as a compact stochastic surrogate for morphology-aware neurite forecasting.
MGRD combines multi-scale centerline conditioning, four spatial gates, joint residual-sequence denoising, and zero-initialized morphology adaptation.
It improves trajectory-wise mean performance on iPSC-derived neuron microscopy, while the controlled simulation comparison isolates the error reduction associated with morphology-gated adaptation.
Across the directional transfer matrix, experimental-to-experimental training reduces MAE relative to synthetic-to-experimental training for both experimental targets and both forecasting models.
Trajectory-wise mean performance also improves image-space and skeleton metrics across 10--40-min mouse cortical-neurosphere sampling intervals, including the 13~h~20~min final horizon.
Repeated samples provide a variance-based case-ranking signal for selective review, while MGRD remains markedly smaller and faster to train than gSTA and has lower cached single-trajectory inference latency.

The results should be interpreted within the scope of the present study.
MGRD operates on two-dimensional, single-channel image crops, and its morphology descriptors reduce each latest observation to a skeleton and distance transform. Illumination or contrast drift, focus changes, microscope-specific noise and preprocessing artifacts, out-of-plane motion, neurites entering the field of view, and abrupt branching, extension, retraction, or fragmentation without visible precursors in the input history therefore remain incompletely represented.
The external evaluation spans four long-duration videos from two specimens without target-domain fine-tuning, which supports transfer across the tested culture and cadence shifts but does not establish performance across laboratories, microscopes, or disease models.
Across the evaluated datasets and sampling intervals, sample variance consistently provides a practical signal for ranking cases by forecast difficulty.

Future work will extend the morphology pathway to three-dimensional and explicitly topology-aware descriptors, incorporate motion and acquisition metadata, and evaluate faster samplers for larger ensembles.
Prospective multi-laboratory studies should test variance-guided acquisition or review decisions before deployment. Future work should calibrate the variance score against observed forecast error and quantify its sensitivity to these specific imaging perturbations and biological changes that are absent from the input history.
Together, these extensions would move MGRD from retrospective image forecasting toward a more complete digital-twin component that couples efficient morphological prediction with transparent, experimentally actionable measures of forecast reliability.

\section*{Code and data availability}
The code, datasets, and derived evaluation outputs used in this study are maintained in a GitHub repository \url{https://github.com/CMU-CBML/NDD_Diffusion}.
The original gSTA framework and datasets are associated with the ``NDD\_ML'' repository: \url{https://github.com/CMU-CBML/NDD_ML}.

\section*{Declaration of competing interest}
The authors declare no known competing financial interests or personal relationships that could have appeared to influence the work reported in this paper.

\section*{Acknowledgement}
This work was supported in part by the US National Science Foundation (NSF: \#1953323 and \#2332084) and the Manufacturing PA Innovation Program.
The authors thank Genesis Omana Suarez, Toshihiko Nambara, and Takahisa Kanekiyo at Mayo Clinic for providing the human iPSC-derived neuron microscopy data.
This work used the Bridges-2 supercomputer at the Pittsburgh Supercomputing Center through allocation ID eng170006p from the Advanced Cyberinfrastructure Coordination Ecosystem: Services \& Support (ACCESS) program, which is supported by the National Science Foundation, United States grants \#2138259, \#2138307, \#2137603, and \#2138296.
Generative AI tools were used to assist in reviewing and editing author-written manuscript text. No AI tool was used to generate or alter data, figures or results.
The authors reviewed the resulting revisions and take full responsibility for the content and conclusions of this work.

\bibliographystyle{elsarticle-num}
\bibliography{reference}
\end{document}